%% file: preprint.tex
\documentclass{article}

\PassOptionsToPackage{numbers,compress}{natbib}

\usepackage[preprint]{neurips_2026}

\usepackage[utf8]{inputenc} % allow utf-8 input
\usepackage[T1]{fontenc}    % use 8-bit T1 fonts
\usepackage{hyperref}       % hyperlinks
\usepackage{url}            % simple URL typesetting
\usepackage{booktabs}       % professional-quality tables
\usepackage{amsfonts}       % blackboard math symbols
\usepackage{nicefrac}       % compact symbols for 1/2, etc.
\usepackage{microtype}      % microtypography
\usepackage{xcolor}         % colors
\usepackage{amsmath}
\usepackage{amssymb}

\usepackage{graphicx}       % for including images
\usepackage{subcaption}     % subfigures
\usepackage{wrapfig}        % wrapped figures
\usepackage{float}          % precise float placement with [H]
\usepackage{makecell}       % compact multi-line table headers
\usepackage{multirow}       % merged table cells
\usepackage{listings}       % code and pseudocode blocks
\usepackage{tabularx}
\usepackage[table]{xcolor}
\usepackage[capitalize,noabbrev]{cleveref}
\usepackage{longtable}
\usepackage{tcolorbox}
\tcbuselibrary{listings, breakable, skins}

\usepackage{float}

\title{Rewarding Structural Conformance of \\Reasoning using Process Mining}

\author{
  Yongjae Lee\thanks{Equal contribution.}\\
  Dept. of Industrial Engineering\\
  Pusan National University\\
  Busan, Republic of Korea \\
  \textit{yongzzai1102@pusan.ac.kr}\\
  \And
  Taekhyun Park\footnotemark[1]\\
  Dept. of Data Science\\
  Pusan National University\\
  Busan, Republic of Korea \\
  \textit{pthpark1@pusan.ac.kr}
  \And
  Sunghyun Sim\\
  Dept. of Data Science\\
  Changwon National University\\
  Changwon, Republic of Korea \\
  \textit{ssh@changwon.ac.kr}
  \And
  Hyerim Bae\thanks{Corresponding author.}\\
  Dept. of Industrial Engineering\\ 
  Pusan National University\\
  Busan, Republic of Korea \\
  \textit{hrbae@pusan.ac.kr}
}

\begin{document}

\maketitle

\begin{abstract}
  Recent advances in sparse reward policy gradient methods have enabled effective reinforcement learning (RL)-based language model post-training. However, for reasoning tasks such as mathematical problem solving, binarized outcome rewards provide limited feedback on intermediate reasoning steps. While some studies have attempted to address this issue by estimating overall reasoning quality, it remains unclear whether these rewards are reliable proxies for the quality of stepwise reasoning. In this study, we consider reasoning as a structured process and propose \textbf{TACReward}, the reward model that can be seamlessly integrated into sparse reward policy gradient methods without additional human annotation costs or architectural modifications. TACReward aggregates stepwise structural deviations between teacher and policy reasoning using process mining techniques, producing a scalar output reward range of $[0, 1]$ to indicate reasoning quality. Experiments on multiple mathematical reasoning benchmarks demonstrate that integrating the TACReward into sparse reward frameworks encourages the policy model to improve the structural quality of reasoning. Consequently, this leads to consistent performance improvements over existing sparse reward frameworks. Our code and checkpoints are publicly available at \href{https://github.com/Thrillcrazyer/TACReward}{GitHub} and \href{https://huggingface.co/Thrillcrazyer/TACReward7B}{HuggingFace}.
\end{abstract}

\input{sections/01-intro.tex}
\newpage
\input{sections/02-Background.tex}
\input{sections/03-Method.tex}
\input{sections/04-Experiment.tex}
\input{sections/05-Conclusion.tex}

{
  \small
  \bibliographystyle{unsrtnat}
  \bibliography{references}
}

%%%%%%%%%%%%%%%%%%%%%%%%%%%%%%%%%%%%%%%%%%%%%%%%%%%%%%%%%%%%
\newpage
\appendix
\input{sections/Appendix.tex}

%%%%%%%%%%%%%%%%%%%%%%%%%%%%%%%%%%%%%%%%%%%%%%%%%%%%%%%%%%%%

% \input{checklist.tex}

\end{document}

%% file: sections/01-intro.tex
\section{Introduction}
\label{sec:intro}

Recent advances in post-training methods based on RL \cite{ouyang2022training} have established a new paradigm for enhancing the alignment and reasoning capabilities of Large Reasoning Models (LRMs)~\cite{zhang2025survey}. Traditional policy gradient methods, such as Proximal Policy Optimization (PPO)~\cite{schulman2017proximal} and its variants, have achieved significant success. However, PPO incurs substantial computational and memory overhead from maintaining a separate critic network and often suffers from training instability when scaled to large reasoning models. To improve efficiency, Reinforced Leave-One-Out (RLOO)~\cite{ahmadian2024back} introduced a leave-one-out baseline to reduce the variance without a dedicated critic. GRPO~\cite{shao2024deepseekmath} and GSPO~\cite{zheng2025group} have shifted this paradigm by adopting sparse reward signals and removing the critic network and the Generalized Advantage Estimation (GAE). 

Despite these advances, sparse reward policy gradient methods remain challenging to apply to reasoning tasks, such as mathematical problem solving due to their dependence on verifiable outcome rewards (i.e., reward sparsity)~\cite{cui2025process}. The reliance on binary and rule-based signals often produces uniform reward signals when responses within a problem group exhibit similar correctness. Consequently, this yields minimal differentiation and weakened learning gradients~\cite{zhang2025grpo}. One potential solution involves providing step-level labels such as Process Reward Models (PRMs)~\cite{lightman2023let}. However, adopting such dense reward within a sparse reward framework is challenging without architectural modifications, potentially diluting the advantage of stability~\cite{sullivan2025grpo}. Furthermore, defining step-level labels is inherently challenging, and human annotation remains costly~\cite{lightman2023let}. 

Recent studies have proposed sparse reward frameworks that consider reasoning quality without annotation~\cite{cui2025process, fan2025posterior, yang2025treerpo, zhan2025exgrpo}, most of which are based on GRPO. However, these methods often rely on an indirect estimation or approximation of the overall reasoning process quality. This makes it challenging to consider the resulting reward signal as indicator of the quality of the individual reasoning steps. Consequently, it remains unclear whether improvements in cumulative rewards reflect genuine gains in logical maturity or mere over-optimization of surrogate metrics. Therefore, while ensuring that step-level logical integrity is crucial for authentic reasoning, an inherent dilemma arises from sparse reward nature. 

We start from the observation that solutions to a reasoning problem can differ substantially at the surface level yet share a common structural skeleton of cognitive activities. In mathematical problem solving, for instance, one solution may proceed by substitution and another by direct computation, but both typically follow a sequence such as \textit{Formulate Strategy} $\rightarrow$ \textit{Apply Formula} $\rightarrow$ \textit{Simplify Expression} $\rightarrow$ \textit{Verify}. This observation suggests that reasoning quality can be characterized structurally, in terms of the sequence and composition of such activities, rather than the specific content of individual reasoning steps.

\begin{figure}[t]
  \includegraphics[width=\linewidth]{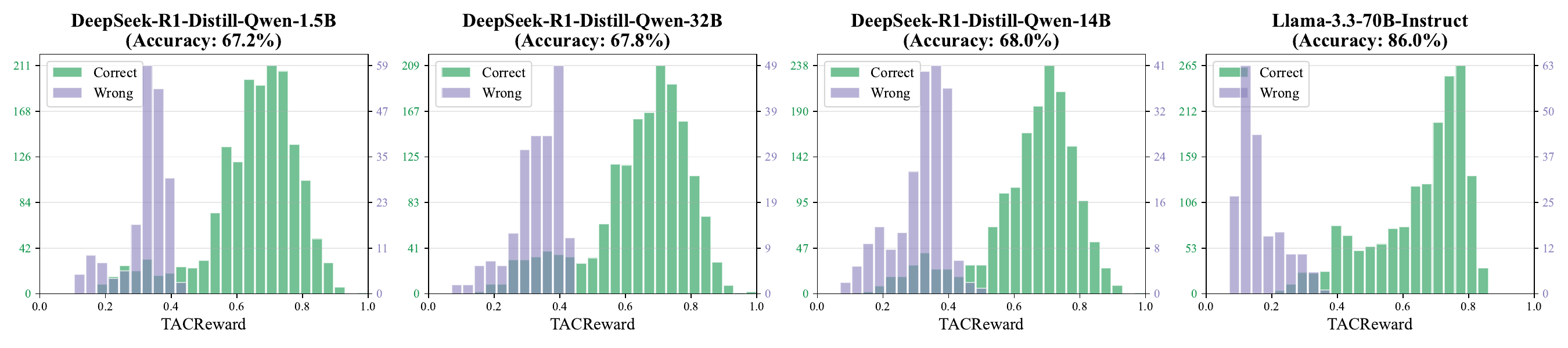}
    \caption{TACReward separates correct from incorrect reasoning. 
    Distributions of TACReward, computed against the reasoning of DeepSeek-R1 (671B)~\cite{guo2025deepseek} as the reference, for correct (green) and incorrect (purple) solutions on DeepMath-103k~\cite{he2025deepmath}. Four models ranging from 1.5B to 70B parameters are evaluated. Correct responses consistently receive a higher degree of alignment than incorrect ones, and the separation is statistically significant for all models (U-test~\cite{mann1947test}, $p < 10^{-10}$).}
    \label{fig:tac_distribution}
  \vspace{-1em}
\end{figure}

Building on this view, we propose \textbf{T}race, \textbf{A}lignment, and \textbf{C}heck \textbf{Reward} (\textbf{TACReward}), a novel reasoning-aware reward model that can be seamlessly integrated into sparse reward policy gradient methods for language model post-training (i.e., the final reward signal is in the range of 0 to 1) with no additional human annotation cost. In TACReward, the reasoning of LRMs is treated as a structured activity sequence (i.e., process) rather than a monolithic output and the policy model is encouraged to improve its degree of alignment with the teacher model's reasoning, which is considered more mature than the policy model. To this end, process mining \cite{van2016data}, a set of techniques specialized in analyzing data recording the execution of processes, is adapted and extended. Our preliminary investigation shows that TACReward, computed against the reference, assigns systematically higher scores to responses with correct final answers than to those with incorrect ones, with a statistically significant separation between the two distributions (Figure~\ref{fig:tac_distribution}).

% TACReward was evaluated using multiple mathematical benchmarks \cite{hendrycks2021measuring, he-etal-2024-olympiadbench, aime, minerva, KorCSATMathCalculus2026, liu2025llmscapablestablereasoning}. The experimental results showed that integrating TACReward into sparse reward policy gradient methods can enhance performance over RLOO, GRPO, and GSPO across all benchmarks. Notably, GSPO + TACReward achieved an average relative accuracy improvement of 89.2 \%.

%% file: sections/02-Background.tex
\section{Preliminaries}
\label{sec:preliminaries}

\subsection{Event Log \& Process Mining}

\begin{wrapfigure}{R}{0.5\textwidth}
    \vspace{-1.2em}
	\centering
	\includegraphics[width=\linewidth]{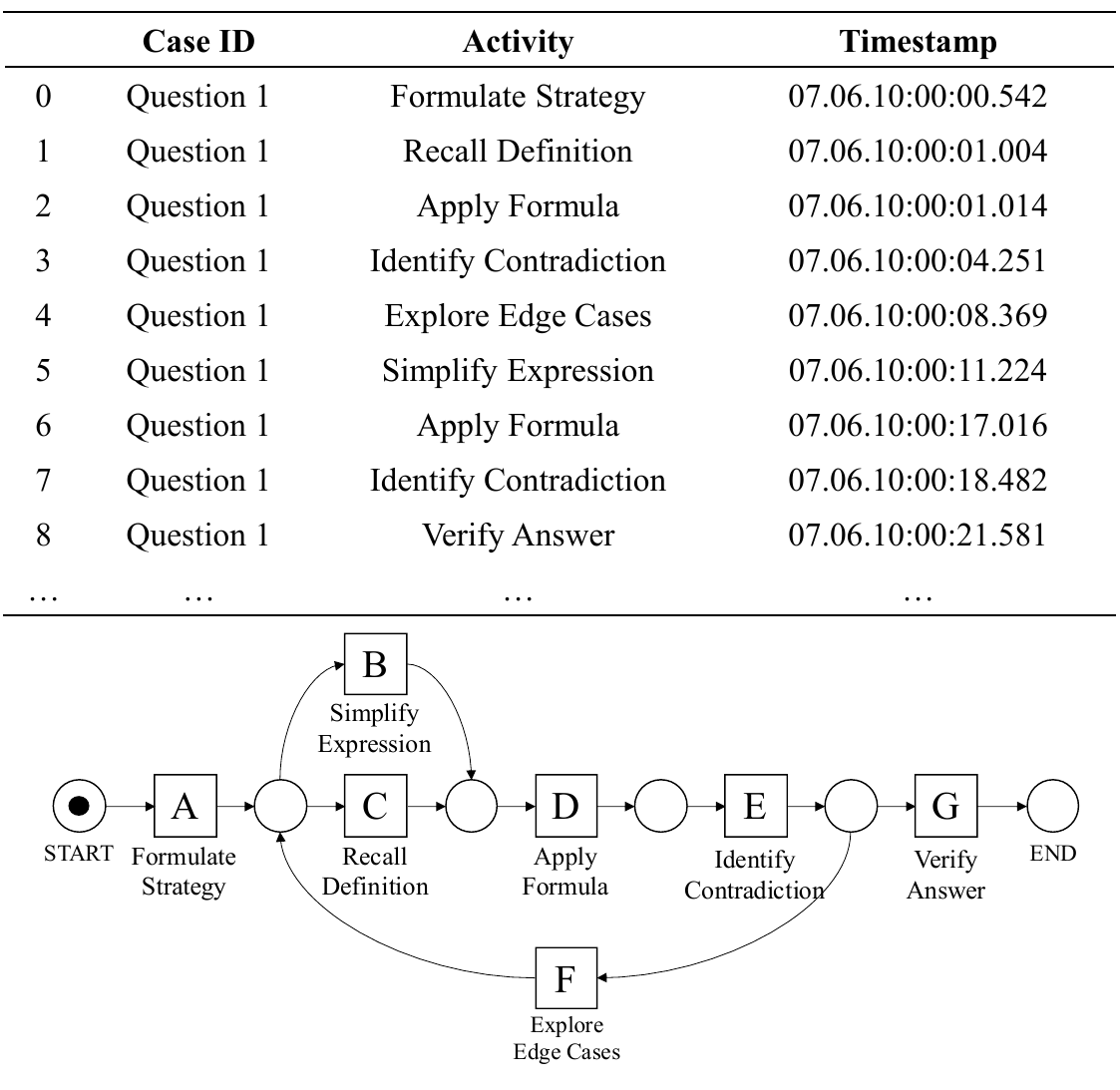}
    \caption{Example of mathematical reasoning event log with a single case id and the corresponding process model inferred from it. The boxes in the map represent activities, and the arrows indicate the flow. The circles denote the places, which are used to model the state of the process.}
    \label{fig:logmap}
    \vspace{-2.5em}
\end{wrapfigure}

An event log \cite{van2011process} is hierarchically structured data that records the execution of a process.
The formal definitions of events, traces, and logs are as follows:
\paragraph{Definition (Event, Trace, and Log).} E is the event universe. An event is a tuple $e=\left(c,a,t,d_{1},\dots,d_{m}\right)\in E$, where $c$ is the case id, $a$ is the activity, $t$ is the timestamp, and $d_{1},\dots,d_{m}$ are the additional attributes. A trace $\sigma = \left\langle e_{1},e_{2},\dots,e_{n}\right\rangle \in E^{*}$ is a finite, non-empty sequence of events such that a trace contains all and only the events in the same case id. An event log $\mathbb{L}$ is a multiset of traces.

Process mining \cite{van2016data} is a set of techniques that uses event logs to discover process models, check the conformance between event log and process model \cite{carmona2022conformance}, and enhance or extend the models. \cref{fig:logmap} presents an example of an event log and the process model derived from it. Various discovery algorithms have been proposed ranging from foundational Alpha Miner \cite{van2011process} to more robust approaches, such as inductive miners~\cite{leemans2013discovering}.

\subsection{Notations}
In this study, an autoregressive LRM as a stochastic policy over sequences parameterized by $\theta$ is denoted by policy $\pi_\theta$.
The teacher model is denoted by $\pi_{\phi}$, where $\phi$ is the set of parameters.
Given an input query $x\sim\mathcal{D}$ sampled from the dataset $\mathcal{D}$, the model generates a response $y=\left\langle y_{1},y_{2},
\dots,y_{\left\lvert y\right\rvert}\right\rangle$.

\subsection{Sparse Reward Policy Gradient Methods for LRMs}
\label{sec:sparse-rl}
The policy gradient methods \cite{sutton1999policy} with sparse rewards maximizes expected reward for the responses generated by the policy LRM $\pi_\theta$:
\begin{equation}
    \begin{aligned}
        \mathcal{J}(\theta) = \mathbb{E}_{x\sim\mathcal{D},y\sim\pi_\theta(\cdot|x)}\left[R(x,y)\right]
    \end{aligned}
\end{equation}
where $R(x,y)$ is the reward function that provides a scalar reward signal based on the quality of the 
response $y$ to query $x$. Using the score function estimator \cite{williams1992simple}, the policy 
gradient can be expressed as
\begin{equation}\label{eq:reinforce}
    \begin{aligned}
        \nabla_\theta \mathcal{J}(\theta)
        = \mathbb{E}\Big[\nabla_\theta \log \pi_\theta(y|x)\,\big(R(x,y)-b(x)\big)\Big]
    \end{aligned}
\end{equation}
where $b(x)$ is a baseline independent of $y$, introduced to reduce the variance of the estimator.
Such a baseline acts as a control variate, subtracting any function independent of $y$, and preserving 
the unbiasedness of the gradient estimator. For autoregressive language models, the log-likelihood factorizes 
over the tokens as follows:
\begin{equation}\label{eq:token_logprob}
    \begin{aligned}
        \log \pi_\theta(y|x) = 
        \sum_{t=1}^{|y|}\log \pi_\theta\big(y_t \mid x, y_{<t}\big)
    \end{aligned}
\end{equation}

It is often convenient to define the (Monte Carlo) advantage estimator \cite{sutton1998reinforcement} as: 
$\hat{A}(x,y) \triangleq R(x,y)-b(x)$, such that \cref{eq:reinforce} can be expressed as $\hat{A}(x,y)$.

\subsection{Group-based Policy Gradient Methods for LRMs}
\label{sec:group}

Recent sparse-reward post-training methods often sample multiple responses for the same query $x$ 
\cite{shao2024deepseekmath, zheng2025group}. For each $x$, we draw a group of $G$ responses
\begin{equation}\label{eq:group_sampling}
    \begin{aligned}
        \{y_{i}\}_{i=1}^{G} \sim \pi_\theta(\cdot \mid x)
    \end{aligned}
\end{equation}
and compute their corresponding outcome rewards $\{r_{i}\}_{i=1}^{G} \triangleq \{R\big(x, y_{i}\big)\}_{i=1}^{G}$,
where $r_{i}$ is typically sparse (e.g., binary correctness) in reasoning tasks.
A common baseline choice in this setting is:
\begin{equation}\label{eq:group_baseline}
    \begin{aligned}
        \hat{A}_{i} = r_{i} - \operatorname{mean}\left(\left\{r_{i}\right\}_{i=1}^{G}\right)
    \end{aligned}
\end{equation} 
% which yields the group-relative advantage $\hat{A}_{i} \triangleq r_{i} - \bar{r}$.
% where $\hat{A}_{i} \approx 0$ for all $i$ if rewards within a group are nearly uniform, 
% leading to diminished learning signals \cite{zhang2025grpo}

%% file: sections/03-Method.tex
\section{Method}
\label{sec:method}

\begin{figure}[ht]
  \begin{center}
    \centerline{\includegraphics[width=\textwidth]{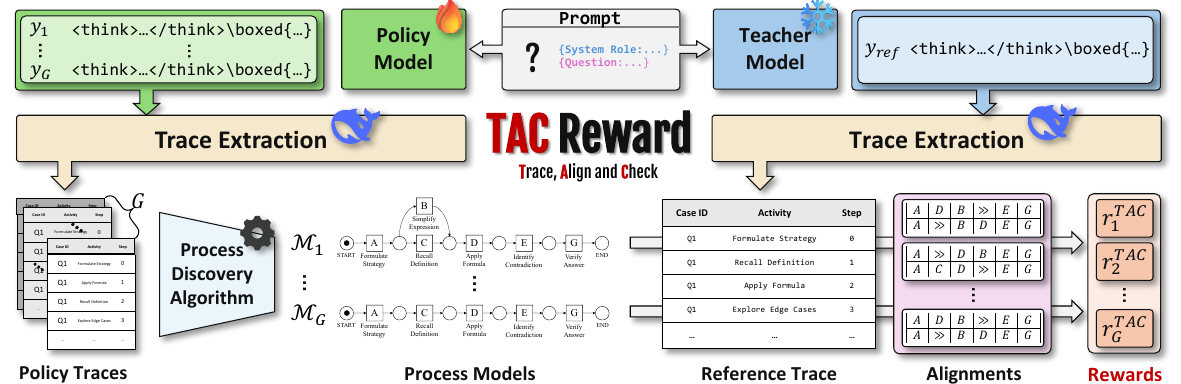}} 
    \caption{
        Overview of the proposed TACReward.
        Given a query $x$, responses $\left\{y_i\right\}_{i=1}^{G}$ are sampled from 
        the policy model $\pi_\theta$ and converts them into policy traces, from which 
        $G$ process models $\left\{\mathcal{M}_i\right\}_{i=1}^{G}$ are discovered. 
        The teacher's reasoning is converted into a reference trace, and each process 
        model $\mathcal{M}_i$ is evaluated against it to obtain the scalar rewards 
        $\left\{r^{TAC}_i\right\}_{i=1}^{G}$. In cases where multiple responses are 
        not generated for the same query, $G=1$.
    }
    \label{fig:TACReward}
  \end{center}
\end{figure}

The proposed TACReward treats the reasoning of an LRM as a structured process rather than a monolithic output and aims to encourage a policy model to improve the degree of alignment with the teacher model’s reasoning process. To this end, techniques of process mining are leveraged and extended. In process mining literature, conformance checking aims to evaluate how a process model can explain an entire event log that contains multiple traces \cite{van2016data}. We extend this paradigm to operate on individual traces. Specifically, TACReward can be categorized into three main steps:
1) \textbf{Trace}: Formalizing the reasoning trace of both the policy and teacher models.
2) \textbf{Align}: Examines the alignment between the policy and reference traces.
3) \textbf{Check}: Evaluate how well the reasoning process of the policy model conforms to that of the teacher model.

The proposed TACReward is illustrated in \cref{fig:TACReward}. 

\subsection{Trace: Formalize Reasoning}

\begin{wraptable}{R}{0.45\textwidth}
  \vspace{-1.2em}
    \centering
    \caption{Taxonomy of 20 mathematical reasoning activities}
    \label{tab:reasoning-activities}
    \footnotesize
    \setlength{\tabcolsep}{5pt}
    \renewcommand{\arraystretch}{1.12}
    \resizebox{\linewidth}{!}{%
    \begin{tabular}{r l r l}
        \toprule
        \textbf{\#} & \textbf{Activity} & \textbf{\#} & \textbf{Activity}\\
        \midrule
        1  & Start Problem               & 11 & Justify Step \\
        2  & Recall Definition           & 12 & Explore Edge Cases \\
        3  & Identify Known Results      & 13 & Identify Contradiction \\
        4  & Formulate Strategy          & 14 & Interpret Result \\
        5  & Apply Known Formula         & 15 & Check Validity \\
        6  & Simplify Expression         & 16 & Verify With Example \\
        7  & Change of Variable          & 17 & Refine or Change Strategy \\
        8  & Evaluate Limit or Integral  & 18 & Conclude Final Result \\
        9  & Perform Comparison          & 19 & Recheck Original Question \\
        10 & Apply Theorem               & 20 & End Problem \\
        \bottomrule
    \end{tabular}%
    }
    \vspace{1.0em}
\end{wraptable}

The computation of the TACReward begins by prompting the policy model to generate a reasoning process for a given problem $x$. For each problem, the model produces $G$ candidate responses, $\{y_{i}\}_{i=1}^{G}$, each containing a sequence of reasoning steps (see \cref{app:system:prompt} for the generation prompt). Subsequently, these responses were formalized into reasoning traces by segmenting them into discrete steps and formatting them to conform to an event log schema. Each structural units of reasoning represent a distinct activity. The taxonomy of the 20 activities is adopted from related studies and mathematical problem-decomposition frameworks~\cite{qin2025decomposing, berti2025configuring, polya2014solve,ritter2019act}. A complete list of the activities is provided in \cref{tab:reasoning-activities}. \cref{tab:taxonomy} provides the ablation study on the taxonomy.

To extract the traces, DeepSeek-V3.2~\cite{liu2025deepseek}, a general-purpose language model, is employed.
Let $\pi_{\psi}$ denote a general-purpose model, the policy traces are formalized as:
\begin{equation}
  \left\{\sigma_{i}\right\}_{i=1}^{G} = \pi_{\psi}\left(\pi_{\theta}(\cdot \mid x)\right)
\end{equation}
where $\sigma_{i}$ is the trace corresponding to the $i$-th response.
Similarly, the reference trace is:
\begin{equation}
  \sigma^{ref} = \pi_{\psi}\left(\pi_{\phi}(\cdot \mid x)\right)
\end{equation}
where the teacher model generates a single response for each problem. 
The prompt used to formalize the trace is provided in \cref{app:trace:prompt}. 

\subsection{Align: Examine the Alignment}

In this step, each policy trace is transformed into a process model using the Inductive Miner (IM) algorithm \cite{leemans2013discovering}. 
Given the policy traces $\{\sigma_i\}_{i=1}^{G}$ and reference trace $\sigma^{ref}$, we discover a process model for each policy trace by 
treating $\{\sigma_i\}$ as a single-trace event log:
\begin{equation}
  \mathcal{M}_{i} = \operatorname{IM}\left(\left\{\sigma_{i}\right\}\right), \quad i=1,\ldots,G
\end{equation}
where $\mathcal{M}_{i}$ denotes the discovered process model corresponding to the $i$th policy trace.

To establish an \textit{alignment} \cite{van2012replaying} between a policy process model $\mathcal{M}_{i}$ and the reference trace 
$\sigma^{ref}$, we relate the \textit{moves} in the model to the \textit{moves} in the trace. Let $\Sigma$ be the set of activity 
labels and let $\gg$ denote a "no-move" (i.e., misalignment) symbol. Given model $\mathcal{M}_{i}$, let $\beta(\mathcal{M}_{i}) 
\subseteq \Sigma^*$ be the set of all possible complete executions sequences in $\mathcal{M}_{i}$.

\paragraph{Definition (Alignment).}
  The alignment of $\sigma^{ref}$ and $\mathcal{M}_i$ is a sequence:
  $\gamma = \langle (a^{ref}_1,a^{m}_1), \dots, (a^{ref}_K,a^{m}_K) \rangle$
  where $(a^{ref}_k,a^{m}_k)\in (\Sigma\cup\{\gg\})\times(\Sigma\cup\{\gg\}) \setminus, 
  \{(\gg,\gg)\}$ such that the top row yields $\sigma^{ref}$ and the bottom row yields 
  $\sigma^\beta \in \beta(\mathcal{M}_i)$. The set of all these alignments is denoted by $\Gamma(\sigma^{ref},
  \mathcal{M}_i)$.

To illustrate this, consider the reference trace $\sigma^{ref} = \langle A, D, B, E, G \rangle$, and the process model in \cref{fig:logmap} as the policy process model $\mathcal{M}_i$. For clarity, two example alignments $\gamma_1$ and $\gamma_2$ are provided in \cref{fig:alignments}. In an alignment, $\gg$ indicates a misalignment (a \textit{log-only} move or \textit{model-only} move), where each column represents a vertical movement. When the same activity is executed in both the reference trace and the model, it is called a \textit{synchronous move}.

\setlength{\intextsep}{0pt}
\begin{wrapfigure}{R}{0.45\textwidth}
\centering
\footnotesize
$\begin{aligned}
  \gamma_{1} &= \left|
  \begin{array}{c|c|c|c|c|c}
    A & D & B & \gg & E & G    \\ \hline
    A & \gg & B & D & E & G
  \end{array}
  \right| \\[8pt]
  \gamma_{2} &= \left|
  \begin{array}{c|c|c|c|c|c}
    A & \gg & D & B & E & G \\ \hline
    A & C & D & \gg & E & G
  \end{array}
  \right|
\end{aligned}$
\caption{
  Two example alignments $\gamma_{1}$ and $\gamma_{2}$. In $\gamma_{1}$, the first move $(A, A)$ is a \textit{synchronous move}. The second move $(D, \gg)$ indicates that the reference trace executes $D$ whereas the model does not. In $\gamma_{2}$, the second move $(\gg, C)$ indicates that the model executes $C$ that is absent from the reference trace.}
\label{fig:alignments}
\vspace{-1.8em}
\end{wrapfigure}

To quantify the discrepancy between the reference traces and the behavior allowed by the policy process model $\mathcal{M}_i$, a finite cost is assigned to each move.

\paragraph{Definition (Cost Function).}
  Let $(a_{ref}, a_m)$ be a move in alignment $\gamma \in \Gamma(\sigma^{ref},
  \mathcal{M}_i)$. The move cost function $\delta_m$ is defined as:
  \begin{equation}
    \delta_m(a_{ref},a_m)=
    \begin{cases}
      0, & a_{ref}=a_m \neq \gg,\\
      w_L(a_{ref}), & a_m=\gg,\\
      w_M(a_m), & a_{ref}=\gg
    \end{cases}
  \end{equation}
where $w_L(\cdot)$ and $w_M(\cdot)$ are non-negative weights and $w_L(\cdot)=w_M(\cdot)=1$ unless stated otherwise.

Finally, the optimal alignment $\gamma_{i}^{*}$ for the $i$-th model is obtained by minimizing the total cost $\delta(\gamma)$, which is the sum of the costs over all moves in the alignment:
\begin{equation}
  \gamma_{i}^{*} 
  = \underset{\gamma \in \Gamma(\sigma^{ref},\mathcal{M}_i)}{\mathrm{argmin}}\,\delta(\gamma)
\end{equation}
\cref{app:additional} provides the computational complexity analysis of finding $\gamma_{i}^{*}$.

\subsection{Check: Evaluate the Conformance}
\label{sec:check}
Through the alignment step, an optimal alignment $\gamma_{i}^{*}$ is established between each policy 
process model $\mathcal{M}_{i}$ and the reference trace $\sigma^{ref}$. Based on this alignment, we 
quantify the \textit{conformance} score $r^{TAC}_{i}$, which serves as the TACReward output for 
each policy response. We compute the conformance by using the F1-score of \textit{fitness} and 
\textit{precision} \cite{van2012replaying}.

\textbf{Fitness.}
Fitness measures how well the process model explains the observed behavior in the 
reference traces \cite{van2016data}. To obtain stable normalization, the finite  
worst-case deviation cost is defined as:
\begin{equation}
  \begin{aligned}
    \delta_{worst}(\sigma^{ref},\mathcal{M}_i)
    = \sum_{e \in \sigma^{ref}} w_L(e)
    &\quad + \min_{\sigma^\beta \in \beta(\mathcal{M}_i)} \sum_{e' \in \sigma^\beta} w_M(e')
  \end{aligned}
\end{equation}
where the cost of $\sigma^{ref}$ with only log moves, plus the minimum cost 
to complete the model using only the model moves. Subsequently, the fitness is computed as
\begin{equation}
  s_{fit}(\sigma^{ref}, \mathcal{M}_{i}) = 1 - \frac{\delta(\gamma_{i}^{*})}{\delta_{worst}(\sigma^{ref},\mathcal{M}_i)}
\end{equation}

\textbf{Precision.}
The precision assesses how often actions that are not present in the trace occur in a process model \cite{van2016data}.
Let $en_M(e)$ be the set of enabled activities in model $\mathcal{M}_{i}$ in the state immediately before event $e$ occurs, 
and let $en_L(e)$ be the set of activities observed in $\sigma^{ref}$ in the same context (i.e., the same prefix immediately 
before $e$). The precision is computed as:
\begin{equation}
  s_{prec}(\sigma^{ref}, \mathcal{M}_{i})
  = \frac{1}{|\mathcal{E}|}\sum_{e \in \mathcal{E}} \frac{|en_L(e)|}{|en_M(e)|},
\end{equation}
where $\mathcal{E}$ denotes the collection of events in the reference trace. A precision value close to 1 indicates that the 
model does not allow significantly more behavior than that observed in the reference trace.

\textbf{Output of TACReward.}
Based on the computed fitness and precision, the output reward $r^{TAC}_{i}$ is:
\begin{equation}
  \label{eq:output}
  r^{TAC}_{i}
  = \frac{2 \cdot s_{fit}(\sigma^{ref}, \mathcal{M}_{i}) \cdot s_{prec}(\sigma^{ref}, \mathcal{M}_{i})}
         {s_{fit}(\sigma^{ref}, \mathcal{M}_{i}) + s_{prec}(\sigma^{ref}, \mathcal{M}_{i})}
\end{equation}
where $r^{TAC}_{i} \in [0, 1]$ denotes the degree of conformance between the reasoning process of the policy model and 
that of the teacher model for the $i$th response. TACReward can be seamlessly integrated into sparse-reward policy 
gradient methods by directly adding its output as the original task reward. Detailed integration procedures are provided 
in \cref{sec:sparse-tac}.

%% file: sections/04-Experiment.tex
\section{Experiments}
\label{sec:experiments}

In this section, we present the experimental results of integrating TACReward into sparse reward policy gradient methods. The details of experiments are described in \cref{app:implementation}.

\subsection{Foundational Policy Gradients Methods with TACReward}

\begin{table}[h]
    \caption{Optimization behavior of sparse-reward RL methods with TACReward (240 training steps)}
    \label{res:withandwithout}
    \vspace*{0.75em}    
    \centering
    \setlength\tabcolsep{2pt} 
    \footnotesize
    {\renewcommand{\tabularxcolumn}[1]{m{#1}}
    \begin{tabularx}{\textwidth}{l*{8}{>{\centering\arraybackslash}X}}
        \toprule
         & 
        \shortstack{\textbf{MATH}\\\textbf{500}} & \textbf{MINERVA} & \textbf{Olympiad} & \shortstack{\textbf{AIME}\\\textbf{2024}} & \shortstack{\textbf{AIME}\\\textbf{2025}} & \shortstack{\textbf{KSAT}\\\textbf{2025}} & \shortstack{\textbf{LiveMath}\\\textbf{Bench}} & \textbf{Avg.*} \\
        \midrule
        \multicolumn{9}{c}{\textit{Qwen2.5-7B-Instruct}} \\
        \midrule
        Baseline (SFT)           & 63.4 & 19.9 & 33.3 & 10.0 & 10.0 & 26.7 & 7.0 & 24.3 \\
        \midrule
        PPO             & 43.2 & 25.4 & 35.6 & 10.0 & 6.7 & 36.7 & 7.0 & 22.3 \\
        PPO + TAC     & \cellcolor{green!15}49.4 & \cellcolor{red!15}24.6 & \cellcolor{green!15}37.6 & \cellcolor{green!15}13.3 & \cellcolor{gray!15}6.7 & \cellcolor{red!15}23.3 & \cellcolor{green!15}12.0 & \cellcolor{red!15}\shortstack{20.8 \scriptsize{(-6.5\%)}}\\
        \specialrule{0.2pt}{1.5pt}{1.5pt}
        RLOO            & 57.4 & 25.4 & 37.6 & 10.0 & 0.0 & 30.0 & 9.0 & 20.4 \\
        RLOO + TAC     & \cellcolor{red!15}57.2 & \cellcolor{red!15}22.8 & \cellcolor{red!15}35.4 & \cellcolor{green!15}13.3 & \cellcolor{green!15}10.0 & \cellcolor{gray!15}30.0 & \cellcolor{green!15}10.0 & \cellcolor{green!15}\shortstack{21.6 \scriptsize{(+6.1\%)}}\\
        \specialrule{0.2pt}{1.5pt}{1.5pt}
        GRPO            & 57.4 & 26.1 & 33.3 & 13.3 & 3.3 & 30.0 & 7.0 & 19.9 \\
        GRPO + TAC     & \cellcolor{green!15}63.4 & \cellcolor{green!15}28.7 & \cellcolor{green!15}38.1 & \cellcolor{gray!15}13.3 & \cellcolor{gray!15}3.3 & \cellcolor{green!15}33.3 & \cellcolor{green!15}9.0 & \cellcolor{green!15}\shortstack{22.5 \scriptsize{(+12.7\%)}}\\
        \specialrule{0.2pt}{1.5pt}{1.5pt}
        GSPO            & 61.4 & 19.9 & 33.3 & 10.0 & 10.0 & 16.7 & 6.0 & 17.2 \\
        GSPO + TAC     & \cellcolor{green!15}64.2 & \cellcolor{green!15}37.1 & \cellcolor{green!15}38.2 & \cellcolor{gray!15}10.0 & \cellcolor{gray!15}10.0 & \cellcolor{green!15}53.3 & \cellcolor{green!15}15.0 & \cellcolor{green!15}\shortstack{32.5 \scriptsize{(+89.2\%)}}\\
        \bottomrule
    \end{tabularx}
    }
\end{table}
\vspace{1.5em}

Table~\ref{res:withandwithout} compares four representative sparse-reward policy optimization methods—PPO, RLOO, GRPO, and GSPO, with and without TACReward-under a fixed training budget of 240 optimization steps on Qwen2.5-7B-Instruct. This setting characterizes early stage optimization rather than converged performance. To better assess the performance, we compute the average score (Avg. *) by \textbf{excluding MATH500 and AIME 2024} and instead focus on the for the remaining five benchmarks.

\paragraph{Strong Synergy with GSPO.} The most significant improvement was observed when TACReward was combined 
with GSPO. Although GSPO alone underperforms in this early stage regime, GSPO + TAC increases Avg. * from 17.2 to 
32.5 (+89.2\%). We attribute this synergy to a \emph{granularity match} between the optimization objective and 
reward signal. GSPO defines the importance ratio and updates it at the \emph{sequence level}, effectively treating 
each sampled response as a single optimization unit. TACReward aggregates the \emph{step-level deviations} revealed 
by the trace alignment into a single \emph{sequence-level} conformance score that represents the reasoning quality of 
one response. This allows TACReward to provide more informative and well-aligned training signal for GSPO 
and can be exploited more effectively than in token-level or step-mismatched objectives.

\paragraph{Compatibility with PPO, RLOO, and GRPO.} TACReward provides consistent improvements for RLOO and GRPO, with relative gains of +6.1\% and +12.7\%, respectively, in Avg.*, suggesting broad compatibility with variance-reduced and group-based objectives. In contrast, adding TACReward to PPO yields limited gains and slightly reduces the overall average score despite improvements on some benchmarks. We attribute this asymmetry to PPO's reliance on dense reward signals: under the sparse reward regime considered in this study, PPO becomes more sensitive to reward variance and credit assignment, whereas RLOO and GRPO are inherently designed for variance reduction and thus absorb the additional signal from TACReward more effectively. \cref{vis:tacnotac} provides visualizations of reasoning processes with and without TACReward during GSPO training.

\subsection{Comparison with State-of-the-Art Models}

\begin{table}[h]
    \caption{Comparison with baseline RL methods using GSPO + TACReward (3000 training steps)}
    \label{res:instruct}
    \vspace*{0.75em}    
    \centering
    \setlength\tabcolsep{3.5pt}
    \footnotesize
    {\renewcommand{\tabularxcolumn}[1]{m{#1}}
    \begin{tabularx}{\textwidth}{l*{7}{>{\centering\arraybackslash}X}}
        \toprule
         & \shortstack{
            \textbf{MATH}\\\textbf{500}} & \textbf{MINERVA} & \textbf{Olympiad} & \shortstack{\textbf{AIME}\\\textbf{2024}} & \shortstack{\textbf{AIME}\\\textbf{2025}} & \shortstack{\textbf{KSAT}\\\textbf{2025}} & \shortstack{\textbf{LiveMath}\\\textbf{Bench}} \\
        \midrule
        \multicolumn{8}{c}{\textit{DeepSeek-R1-Distill-Qwen-7B}} \\
        \midrule
        BASELINE~\cite{shao2024deepseekmath}    & 80.6 & 40.4 & 53.8 & 43.3 & 30.0 & 66.7 & 13 \\
        \rowcolor{green!15} GSPO + TAC (Ours)          & \textbf{90.4} & \textbf{49.3} & \textbf{67.7} & \textbf{50.0} & \textbf{36.7} & \textbf{83.3} & \textbf{24} \\
        DeepMath~\cite{he2025deepmath}          & 83.4 & 42.6 & 48.7 & 34.2 & 30.0 & 63.3 & 17 \\
        SkyWork~\cite{he2025skywork}            & 87.0 & 43.4 & 55.7 & 46.7 & 36.7 & 66.7 & 12 \\
        GRPO-LEAD~\cite{zhang2025grpo}          & 84.6 & 47.4 & 52.3 & 40.0 & 26.7 & 76.7 & 13 \\
        DRGRPO~\cite{liu2025understanding}      & 80.2 & 10.3 & 41.0 & 40.0 & 6.7  & 66.7 & 21 \\
        ExGRPO~\cite{zhan2025exgrpo}            & 82.8 & 41.0 & 47.9 & 23.3 & 16.7 & 63.3 & 13 \\
        Eurus-2~\cite{cui2025process}           & 79.2 & 38.6 & 42.1 & 26.7 & 16.7 & 20.0 & 2 \\
        VeriThinker~\cite{chen2025verithinker}  & 80.2 & 34.9 & 46.4 & 30.0 & 13.3 & 76.7 & 14 \\       
        \midrule
        \multicolumn{8}{c}{\textit{DeepSeek-R1-Distill-Qwen-1.5B}} \\
        \midrule
        BASELINE~\cite{shao2024deepseekmath}    & 69.4 & 26.5 & 40.0 & \textbf{23.3} & \textbf{23.3} & 60.0 & 11 \\
        \rowcolor{green!15} GSPO + TAC (Ours)          & \textbf{74.0} & 27.2 & \textbf{44.0} & 20.0 & 20.0 & \textbf{63.3} & \textbf{14} \\
        DRA-GRPO~\cite{chen2025dra}             & 71.0 & 26.1 & 40.4 & 20.0 & \textbf{23.3} & 43.3 & 5 \\
        Open-RS3~\cite{dang2025reinforcement}   & 69.8 & 26.8 & 39.7 & 20.0 & 20.0 & 50.0 & 6 \\
        STILL-3~\cite{min2024imitate}           & 72.4 & 23.2 & 42.8 & 20.0 & 16.7 & 53.3 & 3 \\
        ExGRPO~\cite{zhan2025exgrpo}            & 71.2 & \textbf{30.9} & 34.4 & 10.0 & 10.0 & 40.0 & 10 \\
        \bottomrule
    \end{tabularx}
    }
\end{table}
\vspace{1.5em}

\begin{figure}[h]
    \centering
    \begin{minipage}[c]{0.45\textwidth}
        \centering
        \includegraphics[width=\linewidth]{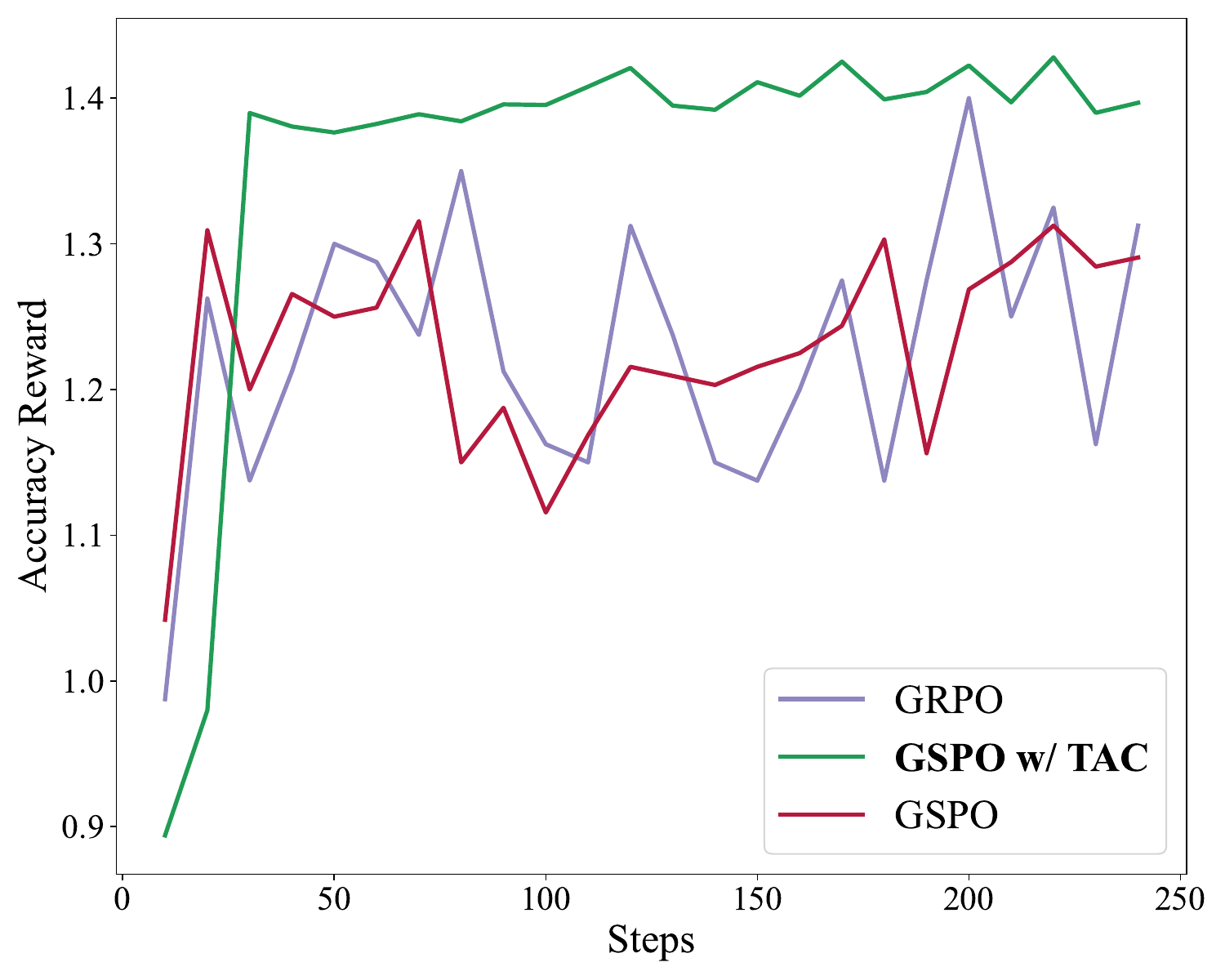}
    \end{minipage}
    \hspace{1em}
    \begin{minipage}[c]{0.45\textwidth}
        \centering
        \includegraphics[width=0.9\linewidth]{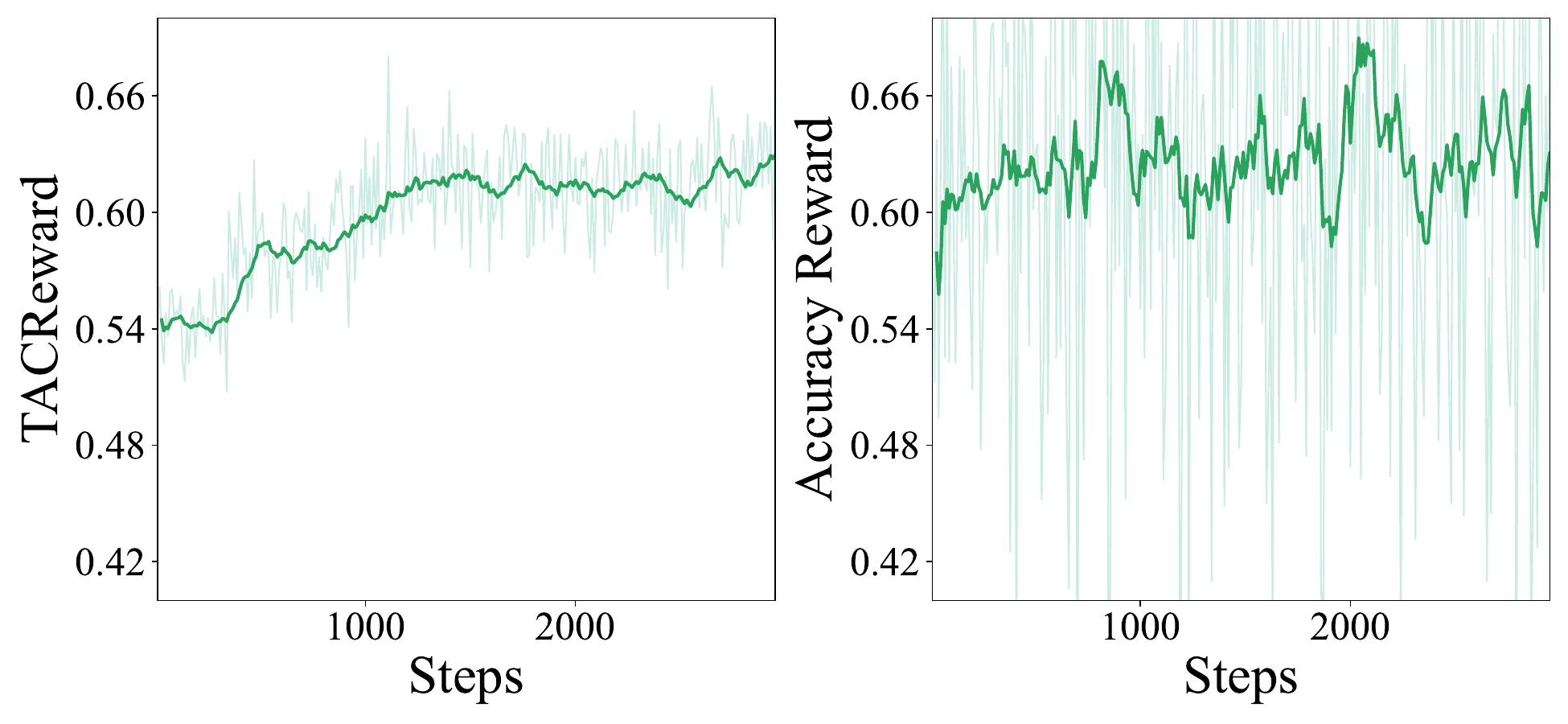} \\[-10pt]
        \includegraphics[width=0.9\linewidth]{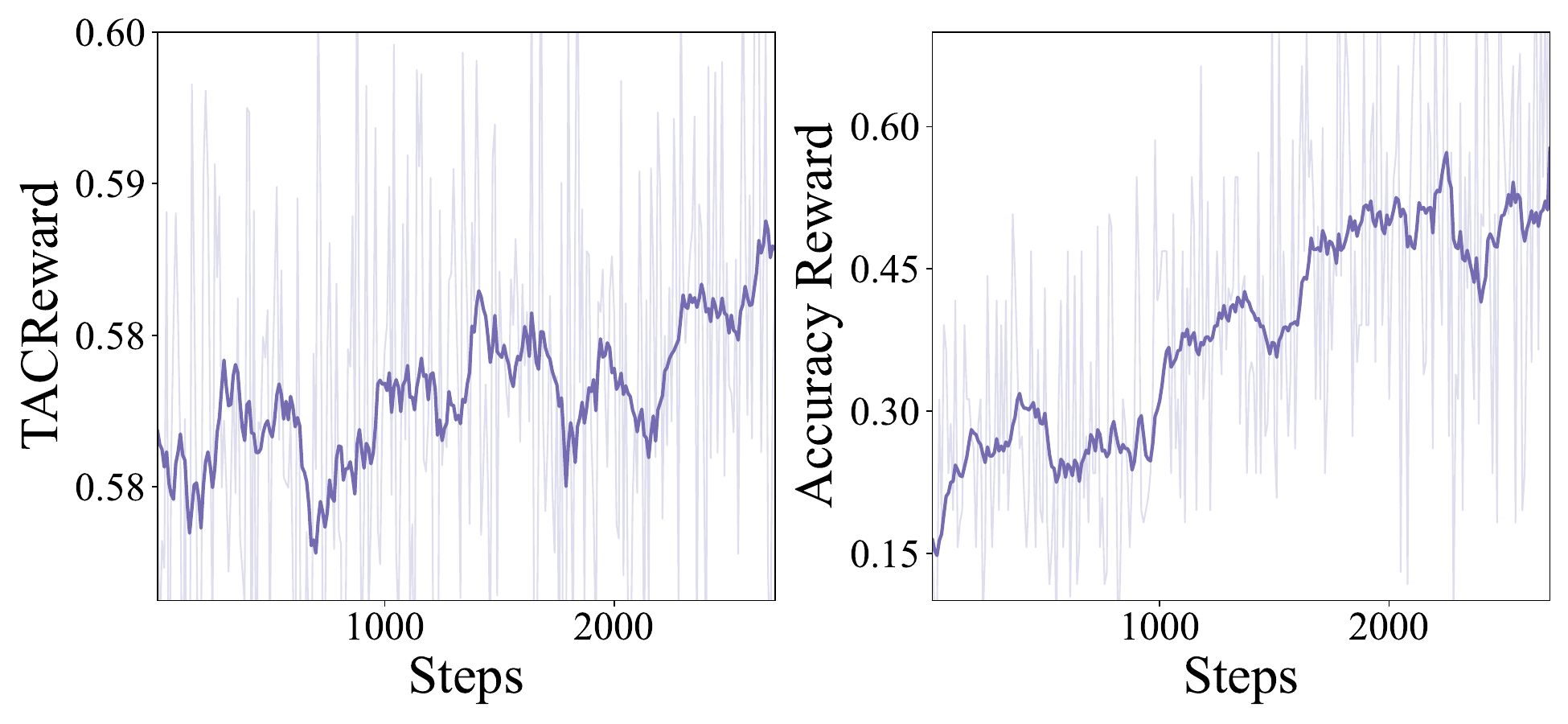}
    \end{minipage}
    \caption{Reward dynamics during GSPO + TAC training. \textbf{Left:} Comparison of accuracy reward among GRPO, GSPO, and GSPO + TAC on the 1.5B policy model. \textbf{Right:} TACReward (left axis) and accuracy reward (right axis) over training steps for the 7B (top) and 1.5B (bottom) policy models, smoothed with a 15-step moving average.}
    \label{fig:TACRewardChange}
\end{figure}
\vspace{1.5em}

Based on the controlled comparison in Table~\ref{res:withandwithout}, we selected GSPO + TACReward as the most 
effective configuration and optimized it further for evaluation. Subsequently, we compared this configuration 
with the recent state-of-the-art RL methods. Table~\ref{res:instruct} compares \textbf{GSPO + TAC} with the 
recent state-of-the-art RL methods for two distilled backbone models of different sizes. Training was conducted 
for 3000 training steps. For both model sizes, GSPO + TAC exhibited strong or state-of-the-art performance across 
the benchmarks. With the 7B backbone, it outperformed all baselines, particularly Olympiad and KSAT 2025, and remained 
competitive (often better) at 1.5B.

\paragraph{Reward Dynamics during Training.} The comparison of reward dynamics among GRPO, GSPO, and GSPO + TAC in the 1.5B policy model in \cref{fig:TACRewardChange} (Left) illustrates that GSPO + TAC achieves stable and high reward trajectories throughout training. Moreover, TACReward steadily increases during training, indicating a progressive better conformance between the policy's reasoning process and the teacher's traces. For both model sizes, the model learns to produce more accurate answers while more closely aligning the reasoning process with the teacher (Right). \cref{vis:training_prog} provides visualizations of reasoning processes as training progresses, 
indicating that the model's reasoning steps became more structured and aligned with those of the teacher over time.

\subsection{Analysis}

\begin{table}[h]
    \caption{Performance comparison of distilled models against the DeepSeek R1 teacher across reasoning benchmarks. Relative degradation from the teacher is shown in parentheses.}
    \label{tab:teacher_size}
    \vspace*{0.75em}    
    \centering
    \setlength\tabcolsep{3.5pt}
    \scriptsize
    {\renewcommand{\tabularxcolumn}[1]{m{#1}}
    \begin{tabularx}{\textwidth}{l c c *{7}{>{\centering\arraybackslash}X}}
        \toprule
        \textbf{Teacher} & \textbf{Size} & \textbf{Teacher Acc.} & \shortstack{\textbf{MATH}\\\textbf{500}} & \textbf{MINERVA} & \textbf{Olympiad} & \shortstack{\textbf{AIME}\\\textbf{2024}} & \shortstack{\textbf{AIME}\\\textbf{2025}} & \shortstack{\textbf{KSAT}\\\textbf{2025}} & \shortstack{\textbf{LiveMath}\\\textbf{Bench}} \\
        \midrule
        \multicolumn{10}{c}{\textit{DeepSeek-R1-Distill-Qwen-1.5B}} \\
        \midrule
        DeepSeek R1          & 685B & 94.17\%          & 49.8 & 19.7 & 20.6 & 0             & 0  & 30    & 12 \\
        R1-Distill-LLaMA     & 70B  & 86\% \tiny{($-8.7\%$)}    & 45.8          & 19.1          & 18.4          & 0             & 0  & 30    & 10  \\
        R1-Distill-Qwen-2.5  & 32B  & 68\% \tiny{($-21.8\%$)}   & 44.8          & 17.6          & 17.3          & 0             & 0  & 26.7  & 12  \\
        R1-Distill-Qwen-2.5  & 14B  & 67.8\% \tiny{($-28.0\%$)} & 44.2          & 15.4          & 17.3          & 3.3  & 0  & 23.3  & 12  \\
        \bottomrule
    \end{tabularx}
    }
\end{table}

\paragraph{Sensitivity to Teacher Quality.} 
A natural concern is whether TACReward depends on access to a frontier-scale teacher. \cref{tab:teacher_size} compares the performance by replacing DeepSeek-R1 (671B) with progressively smaller and less accurate teachers, while keeping all other components fixed. Despite teacher accuracy on DeepMath-103k dropping by up to 28\%, the resulting policy's performance degrades only marginally on most benchmarks. This indicates that TACReward does not require the teacher to be correct on every problem; what matters is the structural pattern of the reasoning trace, which remains informative even when the teacher's final answer is wrong.

\begin{table}[h]
    \caption{Policy performance when varying the trace extractor model.}
    \label{tab:extractor_robustness}
    \vspace*{0.75em}    
    \centering
    \setlength\tabcolsep{3pt}
    \footnotesize
    {\renewcommand{\tabularxcolumn}[1]{m{#1}}
    \begin{tabularx}{\textwidth}{l *{7}{>{\centering\arraybackslash}X}}
    \toprule
    \textbf{Extractor} & \shortstack{\textbf{MATH}\\\textbf{500}} & \textbf{MINERVA} & \textbf{Olympiad} & \shortstack{\textbf{AIME}\\\textbf{2024}} & \shortstack{\textbf{AIME}\\\textbf{2025}} & \shortstack{\textbf{KSAT}\\\textbf{2025}} & \shortstack{\textbf{LiveMath}\\\textbf{Bench}} \\
    \midrule
    DeepSeek-V3.2       & 49.8 & 19.7 & 20.6 & 0.0 & 0.0 & 30.0 & 12 \\
    Gemini-2.5-Flash     & 49.8 & 19.7 & 20.2 & 3.3 & 0.0 & 26.7 & 12 \\
    GPT-5.4-nano         & 50.0 & 20.3 & 20.8 & 3.3 & 0.0 & 30.0 & 12 \\
    \bottomrule
    \end{tabularx}
    }
\end{table}

\paragraph{Sensitivity to Trace Extractor.}
In TACReward, a general-purpose model is utilized to convert raw reasoning responses into activity sequences (\cref{sec:method}). To assess whether this step is bottlenecked by extractor capacity, we replace DeepSeek-V3.2~\cite{liu2025deepseek} with two alternatives spanning different model families: Gemini-2.5-Flash and GPT-5.4-nano. The latter is the smallest member of the GPT-5.4 family. As shown in Table~\ref{tab:extractor_robustness}, all three extractors yield nearly identical policy performance, with GPT-5.4-nano matching or slightly exceeding our default on every benchmark. Trace formalization is thus robust to the choice of extractor as long as the model can follow the 20-activity taxonomy.

\begin{figure}[h]
  \centering
  \includegraphics[width=\textwidth]{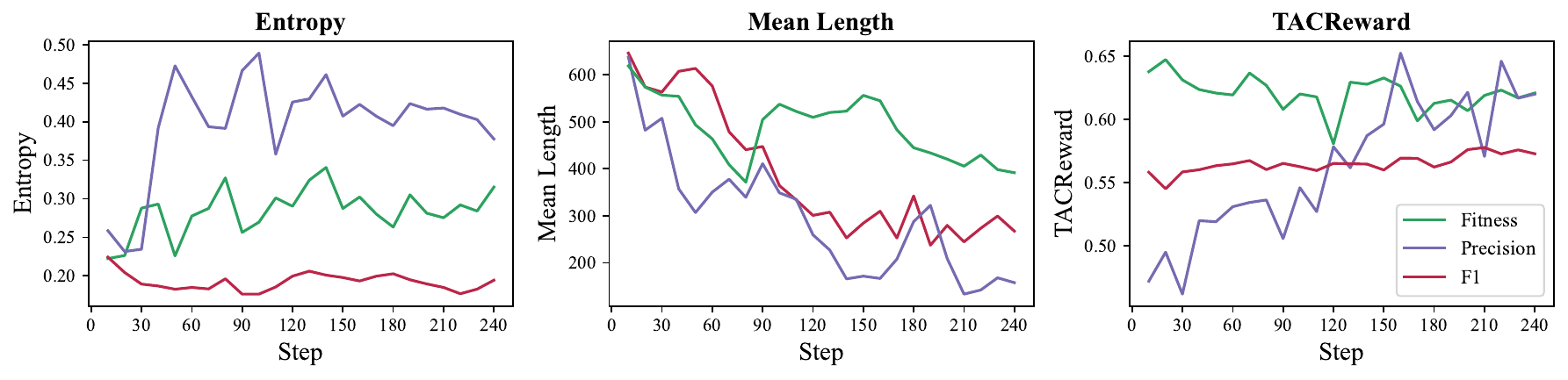}
  \caption{Changes in Entropy, Mean Length, and TACReward for Fitness vs. Precision vs. Both (240 training steps)}
  \label{fig:fitprec}
\end{figure}

\paragraph{Proof of Concept for Using Process Mining.} Based on the definitions in \cite{van2016data}, fitness 
penalizes the missing essential steps, whereas precision penalizes excessive or unjustified behavior. 
Through \cref{fig:fitprec}, we found that \textit{this definition aligns exactly with the expected behavior 
patterns that we want to encourage during training.}

\vspace{1.5em}
\vspace{2em}

When only fitness was used, we expected the model to generate as many reasoning steps as possible and to employ 
a wide variety of activities to avoid penalization for missing essential steps. This is confirmed in \cref{fig:fitprec}, 
where using only fitness (green) results in longer reasoning chains (mean length). However, when only precision is used 
(purple), the model becomes conservative, generating fewer reasoning steps and using fewer activities to minimize excessive 
behavior. This is indicated by the shorter mean length and higher entropy. Both components (red) balance these two tendencies, 
yielding the highest confidence (lowest entropy) and moderate length. \cref{vis:fitprec} visualizes this behavior, and \cref{tab:fitness-precision} confirms that using either component alone degrades performance across benchmarks.

\paragraph{Limitations}
TACReward requires a task-specific taxonomy: the current 20-activity scheme is tailored to mathematical problem solving, and other reasoning domains may demand a new taxonomy together. We further note that our experiments use high-cost API-based teacher and extractor models to showcase the full potential. In practice, one can use smaller models for both roles, and the sensitivity analyses suggest that this would not significantly degrade performance. However, the exact performance trade-offs and optimal configurations when using smaller models remain to be explored.

For additional analyses, \cref{app:additional} provides further results and discussions.

%% file: sections/05-Conclusion.tex
% \section{Related Works}
% \label{sec:relatedworks}

% Recent studies have explored approaches that consider the quality of reasoning processes in a sparse reward 
% policy gradient methods for LRMs \cite{cui2025process, fan2025posterior, yang2025treerpo, zhan2025exgrpo}. 
% However, these methods rely on an indirect estimation or approximation of the overall reasoning quality. 
% For example, metrics such as the final answer correctness, model confidence (e.g., entropy) \cite{zhan2025exgrpo}, 
% preferences for optimized-degraded reasoning pairs \cite{fan2025posterior}, expectations of tree-sampled 
% continuations \cite{yang2025treerpo}, and Process rewards derived from outcome supervision\cite{cui2025process} 
% were used to estimate the overall reasoning quality. Although these methods capture reasoning quality, it 
% remains challenging to consider the resulting reward signal as reflecting the quality of the \textbf{individual} 
% reasoning steps.

\section{Conclusion}
\label{sec:conclusion}
Sparse reward policy gradient methods are challenging to incorporate into reasoning tasks, particularly when learning which reasoning steps to perform and in which order. Moreover, providing such annotations is costly, and their quality can significantly affect training outcomes. In this study, we proposed TACReward, a reasoning-aware reward model that can be seamlessly integrated into sparse reward policy gradient methods for LRM post-training without additional annotations or architectural changes. Instead of teaching the exact intermediate actions, TACReward measures the structural alignment between the reasoning processes of a policy model and that of a more logically mature teacher. The results showed that accounting for structural similarity improves the sparse reward performance and optimization behaviors induced by fitness and precision match their intended effects.

For future works, the underlying mechanism of TACReward is general and we expect it to apply to a broader range of reasoning domains such as code generation. We argue that a particularly promising direction is to extend TACReward to domains where the underlying process is well-defined. In such settings, a teacher model is not strictly necessary: the reference trace can be specified directly from domain knowledge, and TACReward can serve as a structural guardrail that evaluates whether the policy's reasoning conforms to the prescribed procedure. For instance, in loan assessment, the reference trace can encode the standard sequence of steps a credit analyst is expected to perform, and TACReward can guide the policy to internalize a reasoning process appropriate for the task.

%% file: sections/Appendix.tex
\section{Integrating TACReward with Sparse Reward Policy Gradients Methods}
\label{sec:sparse-tac}
In this section, we demonstrate how TACReward can be easily integrated into the foundational sparse reward
policy gradient methods: PPO, RLOO, GRPO, and GSPO. 

\paragraph{Notations}
TACReward is denoted as $\left\{r^{TAC}(x,y_i)\right\}_{i=1}^{G}$.
We set $\left\{r^{acc}(x,y_i)\right\}_{i=1}^{G}$ as the accuracy reward and $\left\{r^{fmt}(x,y_i)\right\}_{i=1}^{G}$ as follows: 
thinking format reward for each response $\left\{y_{i}\right\}_{i=1}^{G}$. The total reward for each query-response pair is denoted as 
$\left\{R(x,y_i)\right\}_{i=1}^{G}$ as described in \cref{sec:preliminaries}. In cases where multiple responses are not generated 
for the same query $G=1$ and are simply denoted as $R(x,y)$.

\subsection{Proximal Policy Optimization (PPO) with TACReward}
\label{app:ppowithTAC}
PPO \cite{schulman2017proximal} generates a single response for each input query (i.e., $G=1$) 
and update the policy model based on the rewards obtained from the response.
For an input $x\sim\mathcal{D}$ and a sampled response $y=\left\langle y_{1},y_{2},
\dots,y_{\left\lvert y\right\rvert}\right\rangle \sim\pi_{\theta_{\mathrm{old}}}(\cdot|x)$,
TACReward can be incorporated into PPO by simply adding it as follows:
\begin{equation}
    \label{eq:total_reward}
    R(x,y) = r^{acc}(x,y) + r^{fmt}(x,y) + {\color{red} r^{TAC}(x,y)}
    -\beta \sum_{t=1}^{\left\lvert y\right\rvert}\bigg(\log \frac{\pi_\theta(y_t|x_t)}{\pi_{\mathrm{ref}}(y_t|x_t)}\bigg),
\end{equation}
where $\pi_{\mathrm{ref}}$ is a fixed reference policy and $\beta>0$ controls the KL regularization strength.
We define a sequence-level advantage with baseline $b(x)$ and broadcast it to all timesteps:
\begin{equation}
    \label{eq:advantage}
    A_t \triangleq A(x,y) = R(x,y) - b(x), \qquad t=1,\dots,\left\lvert y\right\rvert,
\end{equation}
where $b(x)$ is a baseline (e.g., a learned value function $V_\phi(x)$) to reduce variance.
Let $\omega_t(\theta)=\frac{\pi_\theta(y_t|x_t)}{\pi_{\theta_{\mathrm{old}}}(y_t|x_t)}$.
Subsequently, the PPO clipped objective is:
\begin{equation}
    \label{eq:ppo_clip}
    \mathcal{L}_{\mathrm{PPO}}(\theta)
    = \mathbb{E}\!\left[\sum_{t=1}^{\left\lvert y\right\rvert}\min\!\Big(\omega_t(\theta)A_t,\;
    \mathrm{clip}(\omega_t(\theta),1-\epsilon,1+\epsilon)A_t\Big)\right].
\end{equation}

\subsection{REINFORCE with Leave-One-Out (RLOO) with TACReward}
\label{sec:rloowithTAC}
RLOO \cite{ahmadian2024back} generates a group of $G$ responses for each input query (i.e., $G>1$) and updates the policy using a leave-one-out baseline.
Given an input $x\sim\mathcal{D}$ and set of sampled responses
$\{y_i=\langle y_{i,1},\dots,y_{i,|y_i|}\rangle\}_{i=1}^{G}\sim \pi_{\theta_{\mathrm{old}}}(\cdot|x)$,
TACReward can be integrated into RLOO by adding it to the reward for each response as follows:
\begin{equation}
\label{eq:rloo_total_reward}
    R(x,y_i) = r^{acc}(x,y_i) + r^{fmt}(x,y_i) + {\color{red}r^{TAC}(x,y_i)}
    -\beta \sum_{t=1}^{|y_i|}\bigg(\log \frac{\pi_\theta(y_{i,t}|x_{i,t})}{\pi_{\mathrm{ref}}(y_{i,t}|x_{i,t})}\bigg),
    \qquad i=1,\dots,G,
\end{equation}
where $\pi_{\mathrm{ref}}$ is a fixed reference policy and $\beta>0$ controls the KL regularization strength.

RLOO uses a leave-one-out baseline computed from other samples in the group:
\begin{equation}
    \label{eq:rloo_baseline}
    b_i(x) = \frac{1}{G-1}\sum_{j\neq i} R(x,y_j),
    \qquad
    A_i \triangleq A(x,y_i)=R(x,y_i)-b_i(x).
\end{equation}
Using the score function estimator, the RLOO policy gradient is
\begin{equation}
    \label{eq:rloo_grad}
    \nabla_\theta \mathcal{J}(\theta)
    =
    \mathbb{E}\Bigg[
    \frac{1}{G}\sum_{i=1}^{G}
    \Big(\nabla_\theta \log \pi_\theta(y_i|x)\Big)\,A_i
    \Bigg],
    \quad\text{where}\quad
    \log \pi_\theta(y_i|x)=\sum_{t=1}^{|y_i|}\log \pi_\theta(y_{i,t}|x_{i,t}).
\end{equation}

\subsection{Group-based Policy Optimization with TACReward}

GRPO and GSPO bypass the need for the value model by computing the group-relative advantage of each 
response within a response group for the same query. Partially borrowed from \cite{liu2025rethinking}, 
the group-relative advantages for the responses $\left\{y\right\}^{G}_{i=1}$ are computed as:
\begin{equation}
    \begin{aligned}
        \hat{A}_{i,t} = \hat{A}_{i} 
        = \frac{R(x,y_{i})-\operatorname{mean}\left(\left\{R(x,y_{i})_{i=1}^{G}\right\}\right)}
                {\operatorname{std}\left(\left\{R(x,y_{i})_{i=1}^{G}\right\}\right)}
    \end{aligned}
\end{equation}
where all the tokens in $y_{i}$ share the same advantage as $\hat{A}_{i}$.
GRPO then defines the objective as
\begin{equation}
    \begin{aligned}
        \mathcal{J}_{GRPO}(\theta)
        = \mathbb{E}\bigg[\frac{1}{G} \sum_{i=1}^{G} \frac{1}{\left\lvert y_{i}\right\rvert} \sum_{t=1}^{\left\lvert y_{i}\right\rvert}
        \quad\bigg[
            \operatorname{min} \bigg(\omega_{i,t}(\theta) \hat{A}_{i,t}, 
            \operatorname{clip}\left(\omega_{i,t}(\theta), 1-\epsilon,1+\epsilon\right) \hat{A}_{i,t}
            \bigg)\bigg]\bigg]&
    \end{aligned}
\end{equation}
where $\epsilon$ is a hyperparameter for the clipping range, and the importance ratio $\omega_{i,t}(\theta)$ is $\frac{\pi_\theta\left(y_{i,t}|x,y_{i,<t}\right)}{\pi_{\theta_{old}}\left(y_{i,t}|x,y_{i,<t}\right)}$
The GSPO simplifies this objective as follows:
\begin{equation}
    \begin{aligned}
        \mathcal{J}_{GSPO}(\theta) = \mathbb{E}\bigg[\frac{1}{G} \sum_{i=1}^{G}
            \operatorname{min} \bigg(s_{i}(\theta) \hat{A}_{i},
                \operatorname{clip}\left(
                    s_{i}(\theta), 1-\epsilon,1+\epsilon\right) \hat{A}_{i}\bigg)\bigg]&
    \end{aligned}
\end{equation}
where the importance ratio $s_{i}(\theta)$ is $\bigg(\frac{\pi_\theta\left(y_{i}|x\right)}{\pi_{\theta_{old}}\left(y_{i}|x\right)}\bigg)^{\frac{1}{\left\lvert y_{i}\right\rvert}}$
In both GRPO and GSPO, TACReward can be integrated by simply adding it to the reward for each response as follows:
\begin{equation}
    R(x,y_i) = r^{acc}(x,y_i) + r^{fmt}(x,y_i) + {\color{red}r^{TAC}(x,y_i)} \qquad i=1,\dots,G
\end{equation}

\section{Details of Experiments}
\label{app:implementation}

\subsection{Hyperparameters Setting}

We trained the policy model using the TRL framework with DeepSpeed ZeRO-3 optimization. Table~\ref{tab:hyperparameters} 
summarizes the key hyperparameters and distributed training configuration used in experiments~\ref{sec:experiments}. 
All other parameters followed the default settings in TRL GRPOTrainer.

\begin{table}[h]
\centering
\caption{Hyperparameters and Training Configuration for TAC Reward with GSPO}
\label{tab:hyperparameters}
\small
\begin{tabular}{llll}
\toprule
\textbf{Module} & \textbf{Parameter} & \textbf{Value} & \textbf{Description} \\
\midrule
\multirow{4}{*}{Data} 
& \texttt{dataset\_name} & DeepMath-103k & Training dataset. \\
& \texttt{max\_prompt\_length} & 1024 & Maximum input prompt length. \\
& \texttt{max\_completion\_length} & 16384 & Maximum response length. \\
& \texttt{per\_device\_train\_batch\_size} & 2 & Batch size per device. \\
\midrule
\multirow{5}{*}{Model} 
& \texttt{torch\_dtype} & bfloat16 & Model precision. \\
& \texttt{attn\_implementation} & flash\_attention\_2 & Attention implementation. \\
& \texttt{use\_vllm} & True & Enable vLLM for inference. \\
& \texttt{vllm\_mode} & colocate & vLLM execution mode. \\
\midrule
\multirow{4}{*}{Optimizer} 
& \texttt{learning\_rate} & $1 \times 10^{-6}$ & Learning rate for policy optimizer. \\
& \texttt{max\_steps} & 1000 & Total number of training steps. \\
& \texttt{gradient\_accumulation\_steps} & 4 & Gradient accumulation steps. \\
& \texttt{optimizer} & AdamW & Optimizer type (TRL default). \\
\midrule
\multirow{4}{*}{GRPO} 
& \texttt{num\_generations} & 8 & Number of rollouts per prompt. \\
& \texttt{importance\_sampling\_level} & sequence & Importance sampling granularity. \\
& \texttt{beta} & 0.0 & KL penalty coefficient. \\
& \texttt{epsilon} & 0.2 & Clipping parameter (default). \\
\midrule
\multirow{5}{*}{DeepSpeed} 
& \texttt{distributed\_type} & DEEPSPEED & Distributed training backend. \\
& \texttt{zero\_stage} & 3 & ZeRO optimization stage. \\
& \texttt{zero3\_save\_16bit\_model} & True & Save model in 16-bit precision. \\
& \texttt{mixed\_precision} & bf16 & Mixed precision training. \\
& \texttt{num\_processes} & 4 & Number of GPU processes. \\
\bottomrule
\end{tabular}
\end{table}

\subsection{Experimental Settings}

\paragraph{Models \& Baselines} We consider two categories of base models. For non-reasoning base models, 
we used Qwen2.5-1.5B-Instruct and Qwen2.5-7B-Instruct~\cite{qwen2025qwen25technicalreport}, which did not fine-tuned with RL for reasoning capabilities. For the reasoning-enhanced base models, we use DeepSeek-R1-Distill-Qwen-1.5B 
and DeepSeek-R1-Distill-Qwen-7B~\cite{guo2025deepseek}, which already possess reasoning capabilities based on distillation.

\paragraph{Benchmarks \& Evaluation Metric} We train on the DeepMath-103k~\cite{he2025deepmath} dataset and evaluated 
seven challenging mathematical reasoning benchmarks: MATH-500~\cite{hendrycks2021measuring} MINERVA~\cite{minerva}, 
OlympiadBench~\cite{he-etal-2024-olympiadbench}, LiveMathBench~\cite{liu2025llmscapablestablereasoning} and Korean 
CSAT Math Calculus~\cite{KorCSATMathCalculus2026}, and AIME 2024--2025~\cite{aime}. However, we did not address this 
problem MATH500 and AIME 2024 as primary evidence because of potential contamination \cite{wu2025reasoningmemorizationunreliableresults}. 
Therefore, \textbf{the results for MATH-500 and AIME 2024 are reported for reference only}, and the system prompts that was 
previously mentioned, was not used. Instead, all of the main quantitative comparisons and conclusions are obtained from the 
remaining benchmarks. All evaluations used the Pass@1 metric temperature = 0.6, top\_p = 0.95, and max\_tokens = 16384, 
following the flashinfer~\cite{ye2025flashinferefficientcustomizableattention} framework. 
The answers are extracted using the DeepMath Evaluation Library ~\cite{he2025deepmath}.

\paragraph{RL Settings} All experiments are conducted using the TRL framework~\cite{vonwerra2022trl} on 
$4 \times$ NVIDIA H200 GPUs. We employed the DeepSeek API~\cite{guo2025deepseek} as the reward model that 
provides both accuracy and conformance reward signals. We used the AdamW optimizer for optimization at a 
constant learning rate of $1 \times 10^{-6}$. The training was performed with a global batch size of 128, 
utilizing micro-batch sizes of 8 for the 1.5B models and 4 for 7B models, respectively. We set the KL 
divergence coefficient to zero for unconstrained policy updates. The details of hyperparameters are provided 
in \cref{app:implementation}.

\section{Additional Analysis}
\label{app:additional}

\subsection{Computational Complexity of the optimal alignment.}
We analyze the computational cost of TACReward as the length of the reasoning increases. Let $n = |\sigma^{\text{ref}}|$ denote the length of the reference trace, $m = |\sigma_i|$ the length of the $i$-th policy trace, and $|\Sigma|$ the size of the activity set. In our setting, $|\Sigma| \leq 20$ (Section~\ref{sec:method}).

\paragraph{Optimal alignment.}
Computing the optimal alignment $\gamma^*_i$ between a reference trace $\sigma^{\text{ref}}$ and a policy process model $\mathcal{M}_i$ can be cast as an $A^*$ search over the synchronous product of the two~\cite{van2012replaying}. In the general case this search is exponential in the trace length, but three properties of our setting reduce it substantially: (i) the activity taxonomy is fixed, so $|\Sigma|$ is a constant; (ii) the number of reachable intermediate states in each process model is bounded by a constant, since each $\mathcal{M}_i$ is discovered from a single trace; and (iii) the search graph contains no branching beyond local model moves. Under these conditions, the cost of finding $\gamma^*_i$ reduces to $O(n \log n)$.

\paragraph{Output reward.}
Once $\gamma^*_i$ is obtained, the fitness score requires summing local move costs along the alignment, which takes $O(n + m)$. The precision score iterates over events in the reference trace and inspects the set of enabled activities at each step, costing $O(n \cdot |\Sigma|)$. Since $|\Sigma|$ is constant, both scores reduce to $O(n)$, and the harmonic mean in Equation~\ref{eq:output} adds only constant overhead. The overall cost of computing $r_i^{TAC}$ is therefore dominated by the alignment step at $O(n \log n)$.

\subsection{TAC as a Proxy for Step-level Reward}
While $r^{TAC}_{i}$ is delivered as a single scalar value per response, $r^{TAC}_{i}$ is
computed from a fine-grained step-level comparison of reasoning processes via the alignment $\gamma_i^*$. 
Consequently, TACReward provides a proxy for step-level supervision without requiring explicit dense annotations.

The alignment $\gamma_i^*$ decomposes the discrepancy between the policy and teacher reasoning traces into individual 
moves, each incurring local cost $\delta_m(\cdot)$. The total deviation cost performs implicit credit assignment at the 
reasoning transition level, even though the final reward is scalar. A log-only move $(a_{ref}, \gg)$ indicates that the 
policy model omits the reasoning activity that is present in the teacher's trace and corresponds to a missing logical step. 
Conversely, the model-only move $(\gg, a_m)$ reflects redundant or irrelevant reasoning activities 
that are not supported by the teacher's process. 

Under this decomposition, the fitness penalizes missing essential steps, whereas the precision penalizes excessive or unjustified behavior. 
Their harmonic mean captures both completeness and selectivity and prevents premature conclusions and unsupported over generations.

\subsection{Why Single-Trace Can Be Effective.}
Although process discovery is traditionally applied to multitrace event logs, our use case differs 
fundamentally from classical approaches. We aimed to infer a global reasoning process and construct 
a \textbf{local structural abstraction} that captures the permissible reasoning transitions implied 
by a single policy trace. For each single trace, a minimal process model was constructed that could 
be generalized beyond the observed sequence, encoding the ordering constraints, optional branches, 
and loops in the reasoning trace. This abstraction enabled conformance checking, allowing us to penalize 
structurally invalid reasoning behaviors. 

\subsection{Why Response-Level Comparison Is Insufficient.}
A natural question is whether trace formalization can be skipped, and whether reasoning \emph{responses} 
can be directly compared as sequences. Given two token sequences, $y$ and $y'$, the classical edit 
distance~\cite{wagner1974string} is
\begin{equation}
    d_{\mathrm{ed}}(i,j) =
    \min \begin{cases}
    d_{\mathrm{ed}}(i-1,j) + 1, \\
    d_{\mathrm{ed}}(i,j-1) + 1, \\
    d_{\mathrm{ed}}(i-1,j-1) + \mathbb{I}[y_i \neq y'_j]
\end{cases}
\end{equation}
that measures surface-level similarity via insertions, deletions, and substitutions.
However, \(d_{\mathrm{ed}}\) measures only surface edit operations on raw text, and 
cannot reliably capture the structural properties of reasoning (e.g., missing or spurious 
transitions) that are distinct from benign re-orderings \cite{song2024revisiting}. Trace 
formalization separates reasoning structure from linguistic realization, enabling a meaningful 
structural comparison.

\subsection{On Imperfect Reference Traces.}
In this study, our approach does not assume that the reference model is correct. This assumption 
is shared using several existing post training methods, including GRPO and GSPO. These methods 
rely on reward models or KL-divergence regularization with respect to a reference policy, despite 
these references being imperfect \cite{ouyang2022training}. The reference model must be \emph{more 
mature} than the policy being optimized, providing a relatively stable and coherent reasoning process. 
Under this assumption, conformance-based rewards suppress early stage randomness and structurally 
invalid reasoning, whereas outcome-based rewards continue to guide convergence as the training progresses.

\subsection{Role of Reward Components.} TACReward consists of two components: fitness and precision. To investigate the contribution of each component, we conducted an ablation study by training the policy model with only fitness, only precision, and both components. The results are summarized in Table~\ref{tab:fitness-precision}.

\begin{table}[h]
    \caption{Ablation Study on Reward Components: Fitness vs. Precision vs. Both (240 training steps)}
    \label{tab:fitness-precision}
    \centering
    \setlength\tabcolsep{2pt}
    \footnotesize
    {\renewcommand{\tabularxcolumn}[1]{m{#1}}
    \begin{tabularx}{\columnwidth}{l*{5}{>{\centering\arraybackslash}X}}
        \toprule
         &
        \shortstack{\textbf{Minerva}} &
        \shortstack{\textbf{Olympiad}} &
        \shortstack{\textbf{AIME}\\\textbf{2025}} &
        \shortstack{\textbf{KSAT}\\\textbf{2025}} &
        \shortstack{\textbf{LiveMath}\\\textbf{Bench}} \\
        \midrule
        w/ Fit & 30.9 & 30.5 & 6.7 & 43.3 & 10 \\
        w/ Prec & 35.7 & 35.7 & 3.3 & 46.7 & 12 \\
        w/ Both & \textbf{37.1} & \textbf{38.2} & \textbf{10.0} & \textbf{53.3} & \textbf{15} \\
        \bottomrule
    \end{tabularx}} 
\end{table}

\subsection{Role of Mathematical Taxonomy.}
The Mathematical Reasoning Taxonomy (MRT) provides a fixed set of reasoning activities for 
formalizing raw reasoning responses into comparable traces. 

\begin{table}[h]
    \caption{Ablation Study on the Effect of Mathematical Reasoning Taxonomy (MRT) (240 training steps)}
    \label{tab:taxonomy}
    \centering
    \setlength\tabcolsep{2pt}
    \footnotesize
    {\renewcommand{\tabularxcolumn}[1]{m{#1}}
    \begin{tabularx}{\columnwidth}{l*{5}{>{\centering\arraybackslash}X}}
        \toprule
         &
        \shortstack{\textbf{Minerva}} &
        \shortstack{\textbf{Olympiad}} &
        \shortstack{\textbf{AIME}\\\textbf{2025}} &
        \shortstack{\textbf{KSAT}\\\textbf{2025}} &
        \shortstack{\textbf{LiveMath}\\\textbf{Bench}} \\
        \midrule
        w/o MRT & 37.1 & 35.6 & 3.3 & 43.3 & 9 \\
        w/ MRT & \textbf{37.1} & \textbf{38.2} & \textbf{10.0} & \textbf{53.3} & \textbf{15} \\
        \bottomrule
    \end{tabularx}}
\end{table}

Table~\ref{tab:taxonomy} indicates that removing MRT degrades performance, with larger drops on 
multistep benchmarks (e.g., AIME 2025 and KSAT 2025). In this setting, the rewards become less 
sensitive to structural differences and are dominated by surface variation without an explicit taxonomy.

\section{Prompts for Model Training}
\label{sec:app-prompts}

\subsection{Prompt for Generation}
\label{app:system:prompt}

\begin{tcolorbox}[
    title=System Prompt for Generation,
      colback=gray!3,
      colframe=green!60!black,
      fonttitle=\bfseries,
      coltitle=white,
      colbacktitle=green!50!black,
      label={box:reward-system-prompt},
      enhanced,
      breakable,
      sharp corners=south,
      boxrule=0.8pt,
      arc=3mm,
      left=4mm,
      right=4mm,
      top=3mm,
      bottom=3mm
    ]

A conversation between a user and an assistant. The user asks a question, and the assistant solves it. 
The assistant MUST first think through the solution inside $<think>\dots</think>$, and this block MUST contain 
ONLY step-by-step reasoning with no final answer or filler text. Immediately after $</think>$, on a new line, 
the assistant MUST present the final answer exactly once: no duplicate answers, no alternatives, and no rephrasings. 
The entire final result MUST be enclosed in a single \textbackslash boxed\{ \}, and there MUST be no additional 
text outside that single \textbackslash boxed\{ \} in the final answer. Format template: $<think>$Step-by-step 
reasoning goes here.$</think>$ Final answer here with the result in \textbackslash boxed\{ \}.

\vspace{1em}
\rmfamily
\textbf{Example Input Format:}

\vspace{0.5em}

User: This is the problem: \{Question\}\\
Assistant: $<think>$

\end{tcolorbox}

\subsection{Prompt for Formalizing Trace}
\label{app:trace:prompt}

\begin{tcolorbox}[
  title=Prompt Template for Trace Formalization,
  colback=gray!3,
  colframe=blue!60!black,
  fonttitle=\bfseries,
  coltitle=white,
  colbacktitle=blue!50!black,
  label={box:system-prompt},
  enhanced,
  breakable,
  sharp corners=south,
  boxrule=0.8pt,
  arc=3mm,
  left=4mm,
  right=4mm,
  top=2mm,
  bottom=2mm
]

\textbf{\textcolor{blue!70!black}{System Role:}} You are an Expert Mathematical Problem Solver and Process Mining Assistant. Your goal is to convert unstructured logical reasoning steps into a structured, generalized \textbf{CSV event log} ready for process mining analysis.

\vspace{1em}
\textbf{\textcolor{blue!70!black}{Guidelines for Generalization:}}\\
You must select the Activity label that best describes the \textit{intent} of the reasoning step. Use the definitions below to ensure consistency:

\vspace{0.5em}
\begin{minipage}[t]{0.48\linewidth}
\small
\begin{enumerate}
  \item Start Problem
  \item Recall Definition
  \item Identify Known Results
  \item Formulate Strategy
  \item Apply Known Formula
  \item Simplify Expression
  \item Change of Variable
  \item Evaluate Limit or Integral
  \item Perform Comparison
  \item Apply Theorem
\end{enumerate}
\end{minipage}
\hfill
\begin{minipage}[t]{0.48\linewidth}
\small
\begin{enumerate}
  \setcounter{enumi}{10}
  \item Justify Step
  \item Explore Edge Cases
  \item Identify Contradiction
  \item Interpret Result
  \item Check Validity
  \item Verify With Example
  \item Refine or Change Strategy
  \item Conclude Final Result
  \item Recheck Original Question
  \item End Problem
\end{enumerate}
\end{minipage}

\vspace{1em}
\textbf{\textcolor{red!70!black}{Absolute Constraints:}}
\begin{itemize}
  \item NEVER invent new activities. Use ONLY the 20 labels listed above.
  \item ALWAYS select exactly one of the above labels for the Activity column.
  \item Use each activity label according to its abstract logical purpose.
  \item Each step should reflect one unit of reasoning with a concise description.
  \item Process the input line-by-line: output exactly one CSV row per non-empty line.
\end{itemize}

\vspace{1em}
\textbf{\textcolor{blue!70!black}{Example Output Format:}}

\vspace{0.3em}
{\small\ttfamily
\begin{tabular}{@{}llll@{}}
Case ID & Step & Activity & Description \\
\hline
1 & 01 & Start Problem & Problem: Find function g(x) such that... \\
1 & 02 & Formulate Strategy & Break down the problem into constraints \\
1 & 03 & Recall Definition & Remember that $x^n e^{-x}$ is bounded... \\
1 & 04 & Perform Comparison & Compare integrals... \\
1 & 05 & Apply Known Formula & Use integral of $e^{-x}$ from 0 to 1 \\
1 & 06 & Simplify Expression & Solve for c such that total = 1 \\
1 & 07 & Check Validity & Ensure inequality is strict \\
1 & 08 & Conclude Final Result & Final answer: $g(x) = e^{-x} + 1/e$ \\
1 & 09 & End Problem & Problem solved with valid solution \\
\end{tabular}
}

\vspace{1em}
\textbf{\textcolor{blue!70!black}{Input:}}\\
\texttt{\{problem\}}

\vspace{0.5em}
\textit{Return your output \textbf{only} in CSV format starting with the header.}

\end{tcolorbox}

%%%%%%%%%%%%%%%%%%%%%%%%%%%%%%%%%%%%%%%%%%%%%%%%%%%%%%%%%%%%%%%%%%%%%%%%%%%%%%%%%%%%%%%%%%%%%%%%%
%%%%%%%%%%%%%%%%%%%%%%%%%%%%%%%%%%%%%%%%%%%%%%%%%%%%%%%%%%%%%%%%%%%%%%%%%%%%%%%%%%%%%%%%%%%%%%%%%
%%%%%%%%%%%%%%%%%%%%%%%%%%%%%%%%%%%%%%%%%%%%%%%%%%%%%%%%%%%%%%%%%%%%%%%%%%%%%%%%%%%%%%%%%%%%%%%%%
\newpage
\section{Reasoning Process Visualization with and without TACReward}
\label{vis:tacnotac}

\begin{figure}[h]
  \begin{center}
    \centerline{\includegraphics[width=0.75\textwidth]{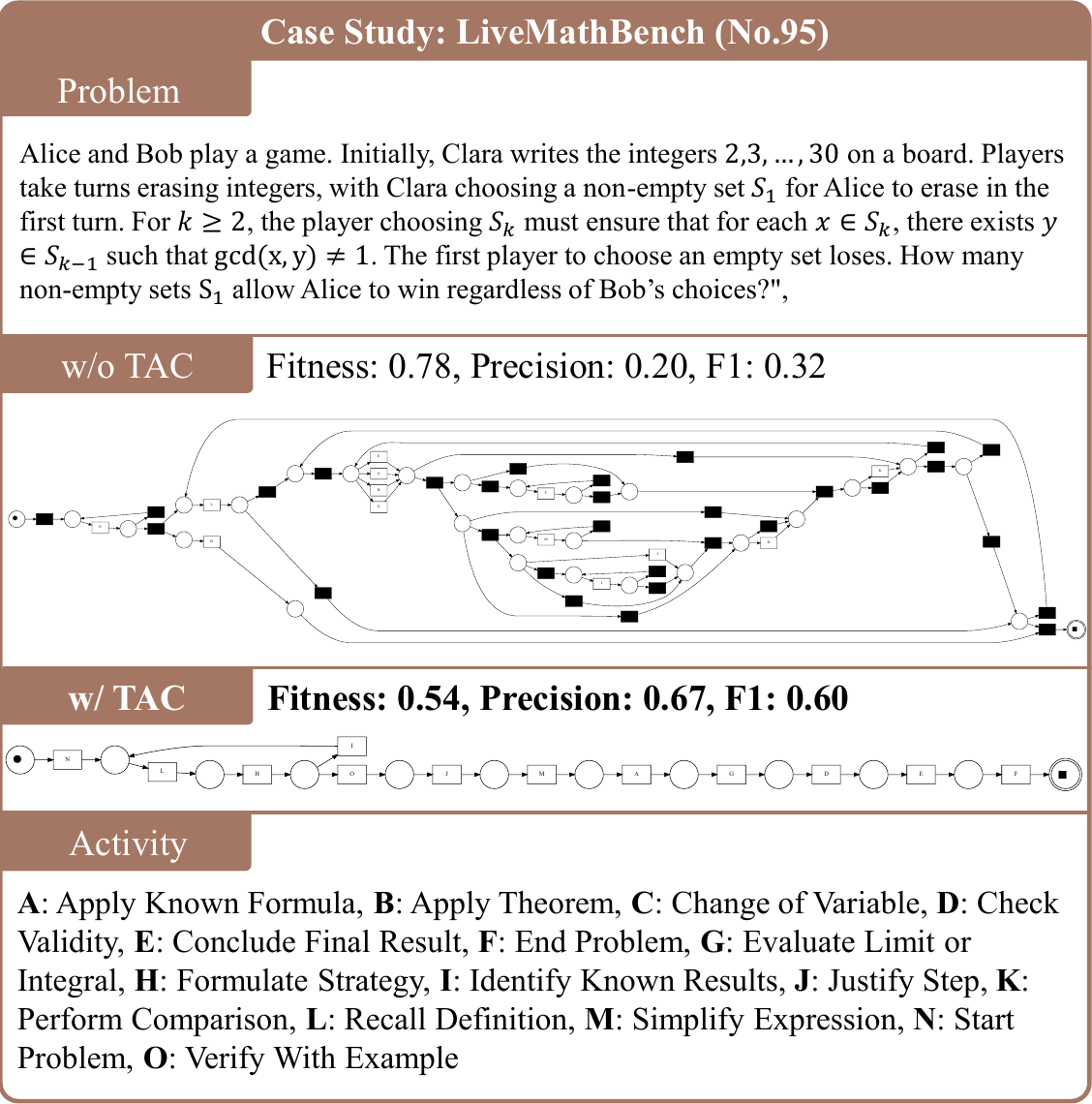}}
  \end{center}
\end{figure}

\newpage
\section{Reasoning Process Visualization across Training Progress}
\label{vis:training_prog}

\begin{figure}[h]
  \begin{center}
    \centerline{\includegraphics[width=0.7\textwidth]{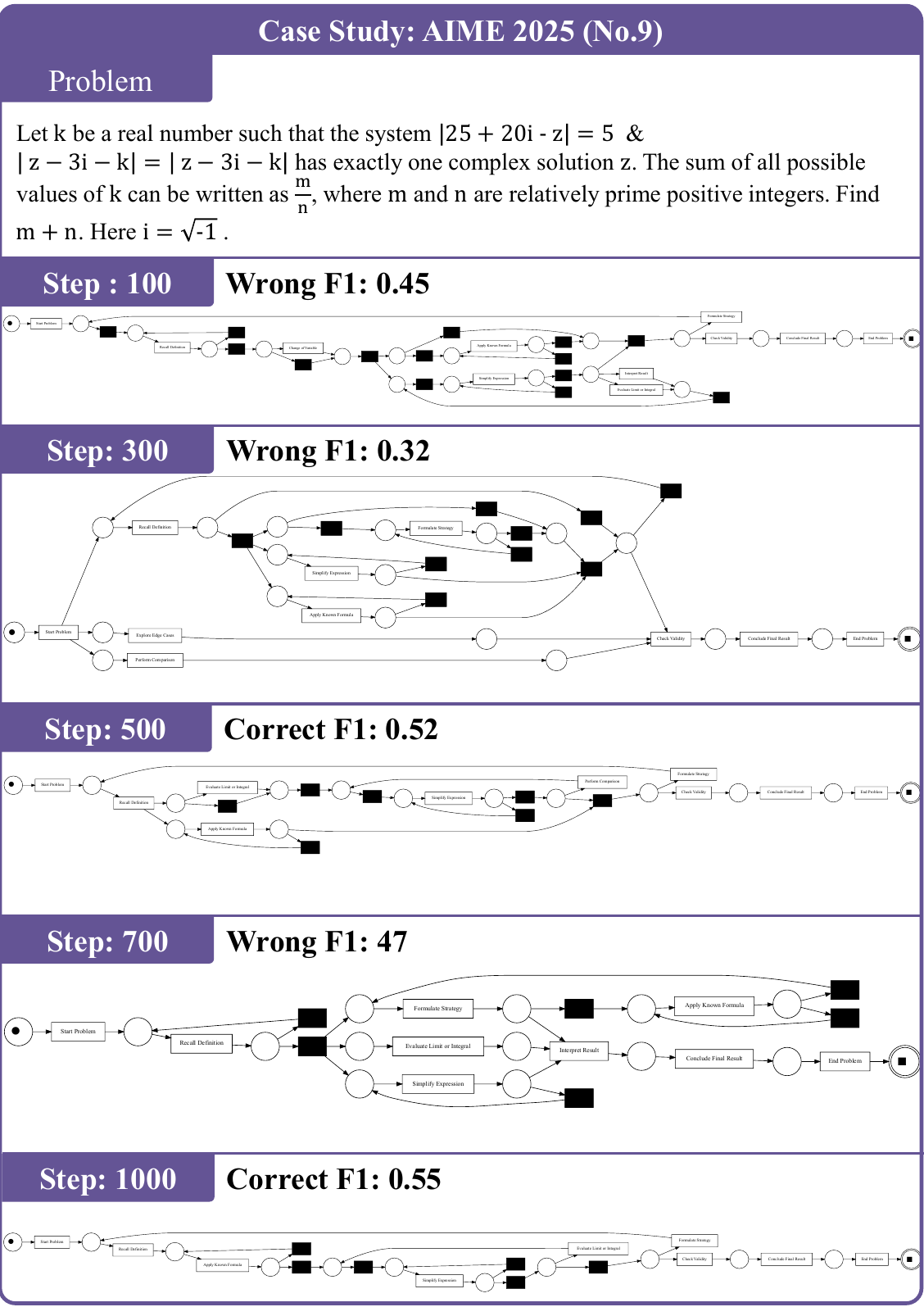}}
  \end{center}
\end{figure}

\section{Reasoning Process Visualization under Fitness, Precision, and F1-Score}
\label{vis:fitprec}

\begin{figure}[H]
  \centering
  \includegraphics[width=0.65\textwidth]{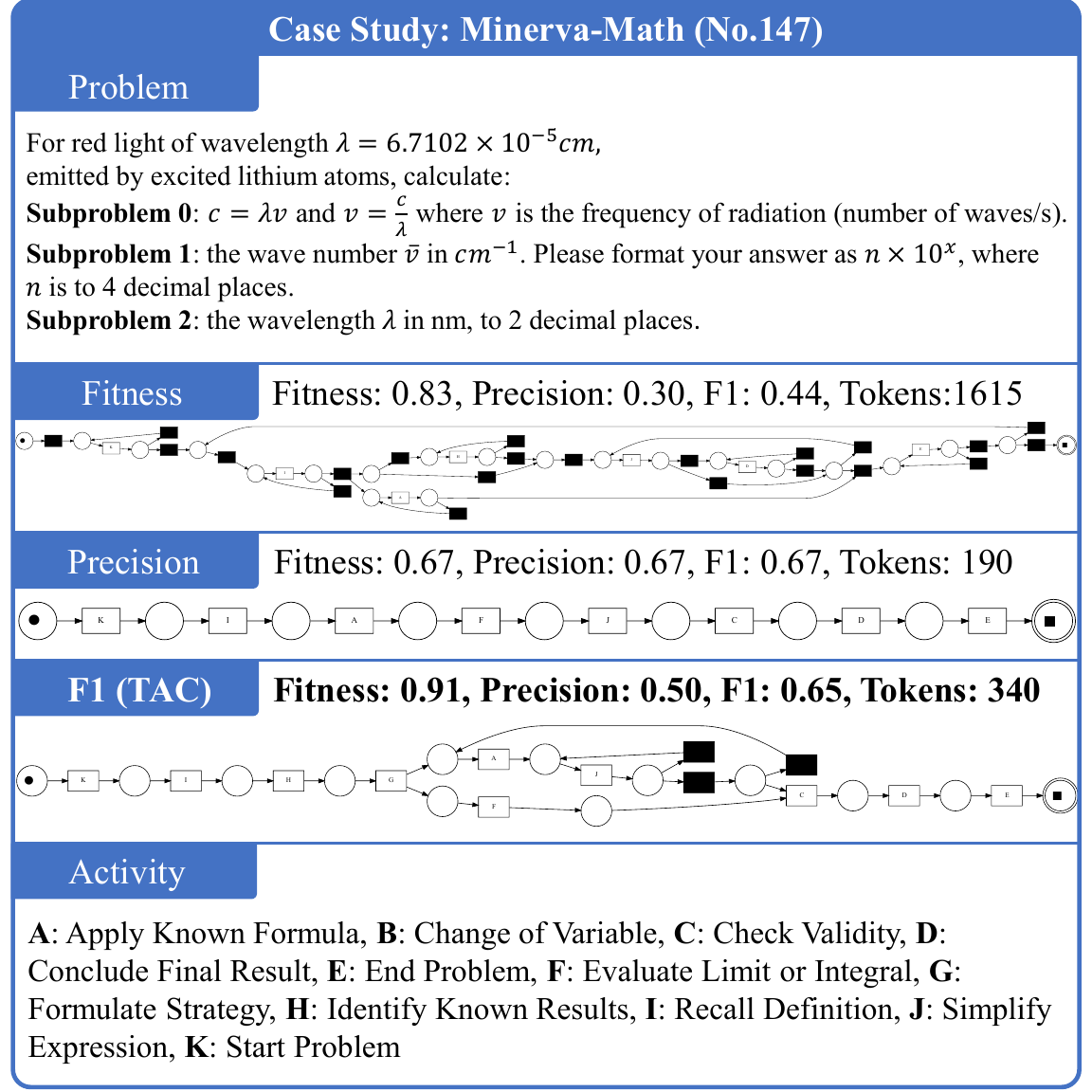}
\end{figure}

\begin{figure}[H]
  \centering
  \includegraphics[width=0.65\textwidth]{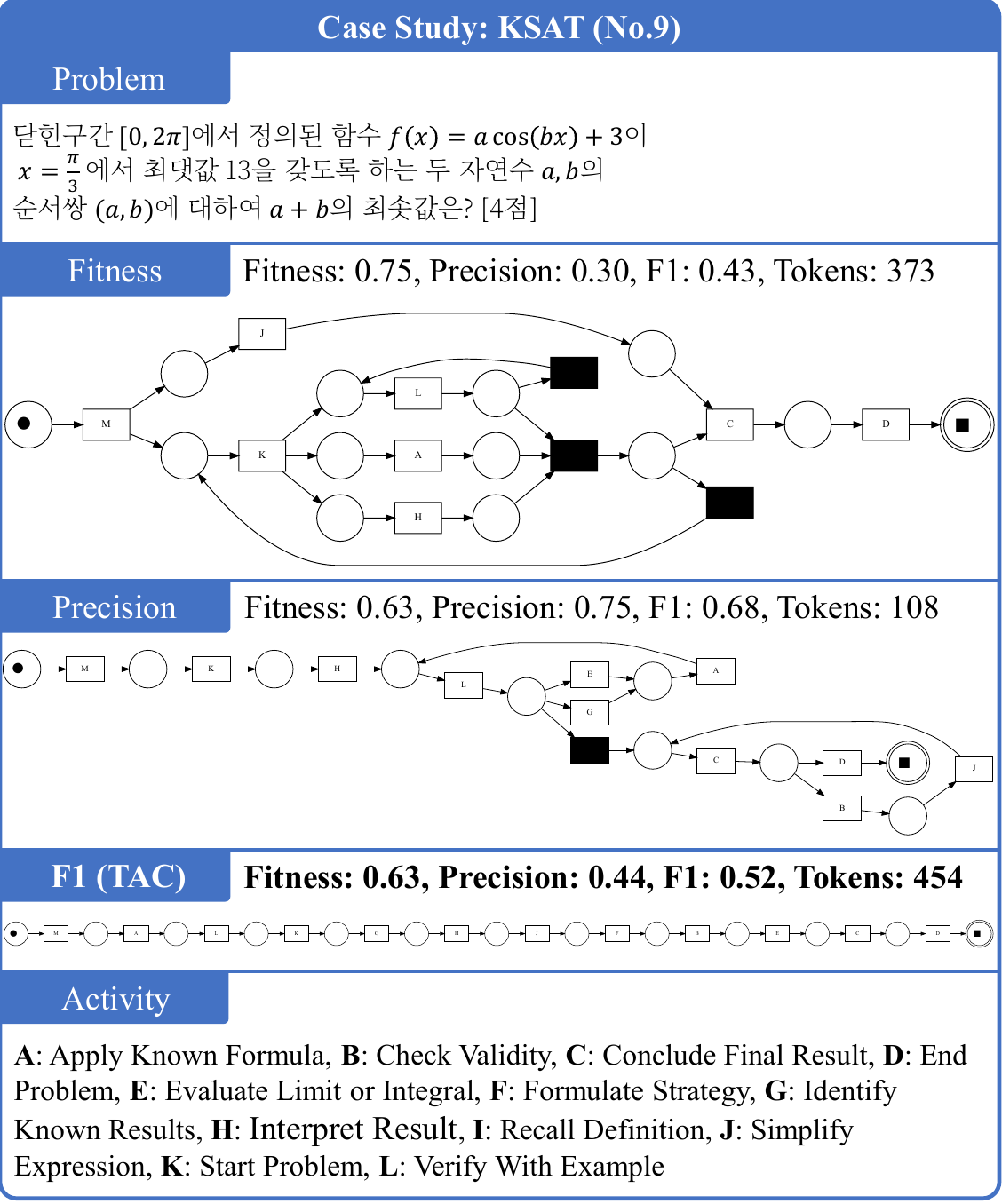}
\end{figure}

\begin{figure}[H]
  \centering
  \includegraphics[width=0.65\textwidth]{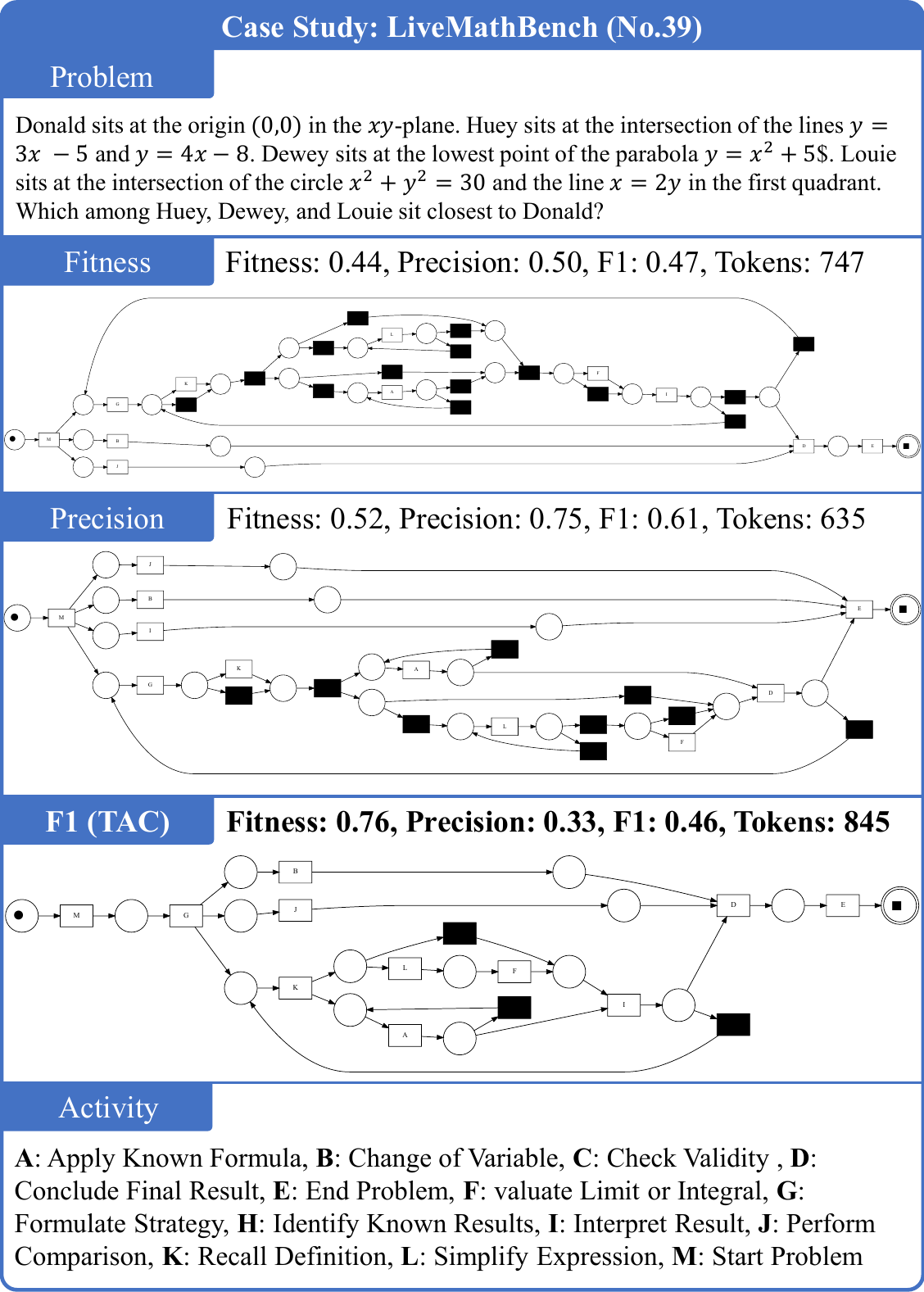}
\end{figure}
\newpage

%% file: references.bib
@article{zhang2025survey,
  title={A survey of reinforcement learning for large reasoning models},
  author={Zhang, Kaiyan and Zuo, Yuxin and He, Bingxiang and Sun, Youbang and Liu, Runze and Jiang, Che and Fan, Yuchen and Tian, Kai and Jia, Guoli and Li, Pengfei and others},
  journal={arXiv preprint arXiv:2509.08827},
  year={2025}
}

@article{schulman2017proximal,
  title={Proximal policy optimization algorithms},
  author={Schulman, John and Wolski, Filip and Dhariwal, Prafulla and Radford, Alec and Klimov, Oleg},
  journal={arXiv preprint arXiv:1707.06347},
  year={2017}
}

@article{shao2024deepseekmath,
  title={Deepseekmath: Pushing the limits of mathematical reasoning in open language models},
  author={Shao, Zhihong and Wang, Peiyi and Zhu, Qihao and Xu, Runxin and Song, Junxiao and Bi, Xiao and Zhang, Haowei and Zhang, Mingchuan and Li, YK and Wu, Yang and others},
  journal={arXiv preprint arXiv:2402.03300},
  year={2024}
}

@article{cui2025process,
  title={Process reinforcement through implicit rewards},
  author={Cui, Ganqu and Yuan, Lifan and Wang, Zefan and Wang, Hanbin and Zhang, Yuchen and Chen, Jiacheng and Li, Wendi and He, Bingxiang and Fan, Yuchen and Yu, Tianyu and others},
  journal={arXiv preprint arXiv:2502.01456},
  year={2025}
}

@article{zhang2025grpo,
  title={Grpo-lead: A difficulty-aware reinforcement learning approach for concise mathematical reasoning in language models},
  author={Zhang, Jixiao and Zuo, Chunsheng},
  journal={arXiv preprint arXiv:2504.09696},
  year={2025}
}

@inproceedings{lightman2023let,
  title={Let's verify step by step},
  author={Lightman, Hunter and Kosaraju, Vineet and Burda, Yuri and Edwards, Harrison and Baker, Bowen and Lee, Teddy and Leike, Jan and Schulman, John and Sutskever, Ilya and Cobbe, Karl},
  booktitle={The Twelfth International Conference on Learning Representations},
  year={2023}
}

@article{sullivan2025grpo,
  title={GRPO is Secretly a Process Reward Model},
  author={Sullivan, Michael},
  journal={arXiv preprint arXiv:2509.21154},
  year={2025}
}

@article{mann1947test,
  title={On a test of whether one of two random variables is stochastically larger than the other},
  author={Mann, Henry B and Whitney, Donald R},
  journal={The annals of mathematical statistics},
  pages={50--60},
  year={1947},
  publisher={JSTOR}
}

@article{fan2025posterior,
  title={Posterior-grpo: Rewarding reasoning processes in code generation},
  author={Fan, Lishui and Zhang, Yu and Chen, Mouxiang and Liu, Zhongxin},
  journal={arXiv preprint arXiv:2508.05170},
  year={2025}
}

@article{yang2025treerpo,
  title={TreeRPO: Tree Relative Policy Optimization},
  author={Yang, Zhicheng and Guo, Zhijiang and Huang, Yinya and Liang, Xiaodan and Wang, Yiwei and Tang, Jing},
  journal={arXiv preprint arXiv:2506.05183},
  year={2025}
}

@incollection{van2016data,
  title={Data science in action},
  author={Van Der Aalst, Wil},
  booktitle={Process mining: Data science in action},
  pages={3--23},
  year={2016},
  publisher={Springer}
}

@article{ahmadian2024back,
  title={Back to basics: Revisiting reinforce style optimization for learning from human feedback in llms},
  author={Ahmadian, Arash and Cremer, Chris and Gall{\'e}, Matthias and Fadaee, Marzieh and Kreutzer, Julia and Pietquin, Olivier and {\"U}st{\"u}n, Ahmet and Hooker, Sara},
  journal={arXiv preprint arXiv:2402.14740},
  year={2024}
}

@article{zheng2025group,
  title={Group sequence policy optimization},
  author={Zheng, Chujie and Liu, Shixuan and Li, Mingze and Chen, Xiong-Hui and Yu, Bowen and Gao, Chang and Dang, Kai and Liu, Yuqiong and Men, Rui and Yang, An and others},
  journal={arXiv preprint arXiv:2507.18071},
  year={2025}
}

@article{liu2025rethinking,
  title={Rethinking GSPO: The Perplexity-Entropy Equivalence},
  author={Liu, Chi},
  journal={arXiv preprint arXiv:2510.23142},
  year={2025}
}

@inproceedings{van2011process,
  title={Process mining manifesto},
  author={Van Der Aalst, Wil and Adriansyah, Arya and De Medeiros, Ana Karla Alves and Arcieri, Franco and Baier, Thomas and Blickle, Tobias and Bose, Jagadeesh Chandra and Van Den Brand, Peter and Brandtjen, Ronald and Buijs, Joos and others},
  booktitle={International conference on business process management},
  pages={169--194},
  year={2011},
  organization={Springer}
}

@incollection{carmona2022conformance,
  title={Conformance checking: foundations, milestones and challenges},
  author={Carmona, Josep and van Dongen, Boudewijn and Weidlich, Matthias},
  booktitle={Process mining handbook},
  pages={155--190},
  year={2022},
  publisher={Springer}
}

@inproceedings{leemans2013discovering,
  title={Discovering block-structured process models from event logs-a constructive approach},
  author={Leemans, Sander JJ and Fahland, Dirk and Van Der Aalst, Wil MP},
  booktitle={International conference on applications and theory of Petri nets and concurrency},
  pages={311--329},
  year={2013},
  organization={Springer}
}

@article{sutton1999policy,
  title={Policy gradient methods for reinforcement learning with function approximation},
  author={Sutton, Richard S and McAllester, David and Singh, Satinder and Mansour, Yishay},
  journal={Advances in neural information processing systems},
  volume={12},
  year={1999}
}

@article{williams1992simple,
  title={Simple statistical gradient-following algorithms for connectionist reinforcement learning},
  author={Williams, Ronald J},
  journal={Machine learning},
  volume={8},
  number={3},
  pages={229--256},
  year={1992},
  publisher={Springer}
}

@book{sutton1998reinforcement,
  title={Reinforcement learning: An introduction},
  author={Sutton, Richard S and Barto, Andrew G and others},
  volume={1},
  number={1},
  year={1998},
  publisher={MIT press Cambridge}
}

@incollection{polya2014solve,
  title={How to solve it: A new aspect of mathematical method},
  author={Polya, George},
  booktitle={How to solve it},
  year={2014},
  publisher={Princeton university press}
}

@article{ritter2019act,
  title={ACT-R: A cognitive architecture for modeling cognition},
  author={Ritter, Frank E and Tehranchi, Farnaz and Oury, Jacob D},
  journal={Wiley Interdisciplinary Reviews: Cognitive Science},
  volume={10},
  number={3},
  pages={e1488},
  year={2019},
  publisher={Wiley Online Library}
}

@article{qin2025decomposing,
  title={Decomposing Elements of Problem Solving: What" Math" Does RL Teach?},
  author={Qin, Tian and Park, Core Francisco and Kwun, Mujin and Walsman, Aaron and Malach, Eran and Anand, Nikhil and Tanaka, Hidenori and Alvarez-Melis, David},
  journal={arXiv preprint arXiv:2505.22756},
  year={2025}
}

@article{berti2025configuring,
  title={Configuring Large Reasoning Models using Process Mining: A Benchmark and a Case Study},
  author={Berti, Alessandro and Kourani, Humam and Park, Gyunam and Van Der Aalst, Wil MP},
  journal={arXiv preprint arXiv:2501.00000},
  year={2025}
}

@article{liu2025deepseek,
  title={Deepseek-v3. 2: Pushing the frontier of open large language models},
  author={Liu, Aixin and Mei, Aoxue and Lin, Bangcai and Xue, Bing and Wang, Bingxuan and Xu, Bingzheng and Wu, Bochao and Zhang, Bowei and Lin, Chaofan and Dong, Chen and others},
  journal={arXiv preprint arXiv:2512.02556},
  year={2025}
}

@article{guo2025deepseek,
  title={Deepseek-r1: Incentivizing reasoning capability in llms via reinforcement learning},
  author={Guo, Daya and Yang, Dejian and Zhang, Haowei and Song, Junxiao and Zhang, Ruoyu and Xu, Runxin and Zhu, Qihao and Ma, Shirong and Wang, Peiyi and Bi, Xiao and others},
  journal={arXiv preprint arXiv:2501.12948},
  year={2025}
}

@article{van2012replaying,
  title={Replaying history on process models for conformance checking and performance analysis},
  author={Van der Aalst, Wil and Adriansyah, Arya and Van Dongen, Boudewijn},
  journal={Wiley Interdisciplinary Reviews: Data Mining and Knowledge Discovery},
  volume={2},
  number={2},
  pages={182--192},
  year={2012},
  publisher={Wiley Online Library}
}

@misc{ye2025flashinferefficientcustomizableattention,
      title={FlashInfer: Efficient and Customizable Attention Engine for LLM Inference Serving}, 
      author={Zihao Ye and Lequn Chen and Ruihang Lai and Wuwei Lin and Yineng Zhang and Stephanie Wang and Tianqi Chen and Baris Kasikci and Vinod Grover and Arvind Krishnamurthy and Luis Ceze},
      year={2025},
      eprint={2501.01005},
      archivePrefix={arXiv},
      primaryClass={cs.DC},
      url={https://arxiv.org/abs/2501.01005}, 
}

@inproceedings{hendrycks2021measuring,
title={Measuring Massive Multitask Language Understanding},
author={Dan Hendrycks and Collin Burns and Steven Basart and Andy Zou and Mantas Mazeika and Dawn Song and Jacob Steinhardt},
booktitle={International Conference on Learning Representations},
year={2021},
url={https://openreview.net/forum?id=d7KBjmI3GmQ}
}

@inproceedings{he-etal-2024-olympiadbench,
    title = "{O}lympiad{B}ench: A Challenging Benchmark for Promoting {AGI} with Olympiad-Level Bilingual Multimodal Scientific Problems",
    author = "He, Chaoqun  and
      Luo, Renjie  and
      Bai, Yuzhuo  and
      Hu, Shengding  and
      Thai, Zhen  and
      Shen, Junhao  and
      Hu, Jinyi  and
      Han, Xu  and
      Huang, Yujie  and
      Zhang, Yuxiang  and
      Liu, Jie  and
      Qi, Lei  and
      Liu, Zhiyuan  and
      Sun, Maosong",
    editor = "Ku, Lun-Wei  and
      Martins, Andre  and
      Srikumar, Vivek",
    booktitle = "Proceedings of the 62nd Annual Meeting of the Association for Computational Linguistics (Volume 1: Long Papers)",
    month = aug,
    year = "2024",
    address = "Bangkok, Thailand",
    publisher = "Association for Computational Linguistics",
    url = "https://aclanthology.org/2024.acl-long.211/",
    doi = "10.18653/v1/2024.acl-long.211",
    pages = "3828--3850"
}

@misc{aime,
  title        = {American Invitational Mathematics Examination ({AIME})},
  organization = {Mathematical Association of America ({MAA})},
  key          = {MAA},
  howpublished = {Mathematics Competition Series},
  url          = {https://maa.org/math-competitions/aime},
  year = {n.d.},
}

@inproceedings{minerva,
 author = {Lewkowycz, Aitor and Andreassen, Anders and Dohan, David and Dyer, Ethan and Michalewski, Henryk and Ramasesh, Vinay and Slone, Ambrose and Anil, Cem and Schlag, Imanol and Gutman-Solo, Theo and Wu, Yuhuai and Neyshabur, Behnam and Gur-Ari, Guy and Misra, Vedant},
 booktitle = {Advances in Neural Information Processing Systems},
 editor = {S. Koyejo and S. Mohamed and A. Agarwal and D. Belgrave and K. Cho and A. Oh},
 pages = {3843--3857},
 publisher = {Curran Associates, Inc.},
 title = {Solving Quantitative Reasoning Problems with Language Models},
 volume = {35},
 year = {2022}
}

@misc{liu2025llmscapablestablereasoning,
      title={Are Your LLMs Capable of Stable Reasoning?}, 
      author={Junnan Liu and Hongwei Liu and Linchen Xiao and Ziyi Wang and Kuikun Liu and Songyang Gao and Wenwei Zhang and Songyang Zhang and Kai Chen},
      year={2025},
      eprint={2412.13147},
      archivePrefix={arXiv},
      primaryClass={cs.AI},
      url={https://arxiv.org/abs/2412.13147}, 
}

@report{KorCSATMathCalculus2026,
  title        = {Mathematics Section (Calculus) of the 2025 Korean College Scholastic Ability Test},
  organization = {Korea Institute for Curriculum and Evaluation(KICE)},
  key          = {KICE},
  year         = {2025},
  institution  = {Korea Institute for Curriculum and Evaluation},
  note         = {Administered on November 14, 2025}
}

@article{he2025deepmath,
  title={Deepmath-103k: A large-scale, challenging, decontaminated, and verifiable mathematical dataset for advancing reasoning},
  author={He, Zhiwei and Liang, Tian and Xu, Jiahao and Liu, Qiuzhi and Chen, Xingyu and Wang, Yue and Song, Linfeng and Yu, Dian and Liang, Zhenwen and Wang, Wenxuan and others},
  journal={arXiv preprint arXiv:2504.11456},
  year={2025}
}

@misc{wu2025reasoningmemorizationunreliableresults,
      title={Reasoning or Memorization? Unreliable Results of Reinforcement Learning Due to Data Contamination}, 
      author={Mingqi Wu and Zhihao Zhang and Qiaole Dong and Zhiheng Xi and Jun Zhao and Senjie Jin and Xiaoran Fan and Yuhao Zhou and Huijie Lv and Ming Zhang and Yanwei Fu and Qin Liu and Songyang Zhang and Qi Zhang},
      year={2025},
      eprint={2507.10532},
      archivePrefix={arXiv},
      primaryClass={cs.LG},
      url={https://arxiv.org/abs/2507.10532}, 
}

@misc{vonwerra2022trl,
  author = {Leandro von Werra and Younes Belkada and Lewis Tunstall and Edward Beeching and Tristan Thrush and Nathan Lambert and Shengyi Huang and Kashif Rasul and Quentin Gallouédec},
  title = {TRL: Transformer Reinforcement Learning},
  year = {2020},
  publisher = {GitHub},
  journal = {GitHub repository},
  howpublished = {\url{https://github.com/huggingface/trl}}
}

@misc{qwen2025qwen25technicalreport,
      title={Qwen2.5 Technical Report}, 
      author={Qwen and : and An Yang and Baosong Yang and Beichen Zhang and Binyuan Hui and Bo Zheng and Bowen Yu and Chengyuan Li and Dayiheng Liu and Fei Huang and Haoran Wei and Huan Lin and Jian Yang and Jianhong Tu and Jianwei Zhang and Jianxin Yang and Jiaxi Yang and Jingren Zhou and Junyang Lin and Kai Dang and Keming Lu and Keqin Bao and Kexin Yang and Le Yu and Mei Li and Mingfeng Xue and Pei Zhang and Qin Zhu and Rui Men and Runji Lin and Tianhao Li and Tianyi Tang and Tingyu Xia and Xingzhang Ren and Xuancheng Ren and Yang Fan and Yang Su and Yichang Zhang and Yu Wan and Yuqiong Liu and Zeyu Cui and Zhenru Zhang and Zihan Qiu},
      year={2025},
      eprint={2412.15115},
      archivePrefix={arXiv},
      primaryClass={cs.CL},
      url={https://arxiv.org/abs/2412.15115}, 
}

@article{he2025skywork,
  title={Skywork open reasoner 1 technical report},
  author={He, Jujie and Liu, Jiacai and Liu, Chris Yuhao and Yan, Rui and Wang, Chaojie and Cheng, Peng and Zhang, Xiaoyu and Zhang, Fuxiang and Xu, Jiacheng and Shen, Wei and others},
  journal={arXiv preprint arXiv:2505.22312},
  year={2025}
}

@article{liu2025understanding,
  title={Understanding r1-zero-like training: A critical perspective},
  author={Liu, Zichen and Chen, Changyu and Li, Wenjun and Qi, Penghui and Pang, Tianyu and Du, Chao and Lee, Wee Sun and Lin, Min},
  journal={arXiv preprint arXiv:2503.20783},
  year={2025}
}

@article{zhan2025exgrpo,
  title={ExGRPO: Learning to reason from experience},
  author={Zhan, Runzhe and Li, Yafu and Wang, Zhi and Qu, Xiaoye and Liu, Dongrui and Shao, Jing and Wong, Derek F and Cheng, Yu},
  journal={arXiv preprint arXiv:2510.02245},
  year={2025}
}

@article{chen2025verithinker,
  title={VeriThinker: Learning to Verify Makes Reasoning Model Efficient},
  author={Chen, Zigeng and Ma, Xinyin and Fang, Gongfan and Yu, Ruonan and Wang, Xinchao},
  journal={arXiv preprint arXiv:2505.17941},
  year={2025}
}

@article{chen2025dra,
  title={Dra-grpo: Exploring diversity-aware reward adjustment for r1-zero-like training of large language models},
  author={Chen, Xiwen and Zhu, Wenhui and Qiu, Peijie and Dong, Xuanzhao and Wang, Hao and Wu, Haiyu and Li, Huayu and Sotiras, Aristeidis and Wang, Yalin and Razi, Abolfazl},
  journal={arXiv preprint arXiv:2505.09655},
  year={2025}
}

@article{dang2025reinforcement,
  title={Reinforcement Learning for Reasoning in Small LLMs: What Works and What Doesn't},
  author={Dang, Quy-Anh and Ngo, Chris},
  journal={arXiv preprint arXiv:2503.16219},
  year={2025}
}

@article{min2024imitate,
  title={Imitate, explore, and self-improve: A reproduction report on slow-thinking reasoning systems},
  author={Min, Yingqian and Chen, Zhipeng and Jiang, Jinhao and Chen, Jie and Deng, Jia and Hu, Yiwen and Tang, Yiru and Wang, Jiapeng and Cheng, Xiaoxue and Song, Huatong and others},
  journal={arXiv preprint arXiv:2412.09413},
  year={2024}
}

@inproceedings{song2024revisiting,
  title={Revisiting code similarity evaluation with abstract syntax tree edit distance},
  author={Song, Yewei and Lothritz, Cedric and Tang, Xunzhu and Bissyand{\'e}, Tegawend{\'e} and Klein, Jacques},
  booktitle={Proceedings of the 62nd Annual Meeting of the Association for Computational Linguistics (Volume 2: Short Papers)},
  pages={38--46},
  year={2024}
}

@article{wagner1974string,
  title={The string-to-string correction problem},
  author={Wagner, Robert A and Fischer, Michael J},
  journal={Journal of the ACM (JACM)},
  volume={21},
  number={1},
  pages={168--173},
  year={1974},
  publisher={ACM New York, NY, USA}
}

@article{ouyang2022training,
  title={Training language models to follow instructions with human feedback},
  author={Ouyang, Long and Wu, Jeffrey and Jiang, Xu and Almeida, Diogo and Wainwright, Carroll and Mishkin, Pamela and Zhang, Chong and Agarwal, Sandhini and Slama, Katarina and Ray, Alex and others},
  journal={Advances in neural information processing systems},
  volume={35},
  pages={27730--27744},
  year={2022}
}
